\PassOptionsToPackage{table}{xcolor}
\documentclass[sigconf]{acmart}

\usepackage{lipsum}
\usepackage{array}
\usepackage{epsfig}
\usepackage{graphicx}
\usepackage{float}
\usepackage{wrapfig}
\usepackage{bm,xspace}
\usepackage{comment}
\usepackage{multirow}
\usepackage{balance}
\usepackage{url}
\usepackage{booktabs}
\usepackage{etoolbox,siunitx}
\usepackage{calc}
\usepackage{pifont,hologo}
\usepackage{adjustbox}
\usepackage{enumitem}
\usepackage{bbding}
\usepackage[normalem]{ulem}
\usepackage{contour}
\usepackage[table]{xcolor}
\usepackage{soul}

\usepackage{fancyhdr}

\definecolor{blue}{HTML}{004bb3}
\definecolor{red}{HTML}{cc1100}
\definecolor{orange}{HTML}{cc7700}
\definecolor{gray}{HTML}{efefef}
\definecolor{darkgreen}{rgb}{0, 0.69, 0.31}
\definecolor{darkgray}{HTML}{757575}

\definecolor{cite}{HTML}{3270b5}
\definecolor{link}{HTML}{b53532}
\definecolor{link}{HTML}{cc1100}
\definecolor{scratch}{HTML}{001219}
\definecolor{pretrain}{HTML}{0A9396}

\definecolor{ourscolor}{HTML}{DCFCF9}
\definecolor{offsetcolor}{HTML}{6424D6}

\newcommand{\scratch}{\textcolor{scratch}{$\mathbf{\circ}$\,}\xspace}
\newcommand{\pretrain}{\textcolor{pretrain}{$\bullet$\,}\xspace}
\newcommand{\suppretrain}{\textcolor{red}{$\bullet$\,}\xspace}

\contourlength{0.8pt}

\newcommand{\figref}[1]{Fig.~\ref{#1}}
\newcommand{\tabref}[1]{Tab.~\ref{#1}}
\newcommand{\secref}[1]{Sec.~\ref{#1}}
\renewcommand{\eqref}[1]{Eq.~\ref{#1}}

\newcolumntype{x}[1]{>{\centering\arraybackslash}p{#1}}
\newcolumntype{y}[1]{>{\raggedright\arraybackslash}p{#1}}
\newcolumntype{z}[1]{>{\raggedleft\arraybackslash}p{#1}}
\newcommand{\tablestyle}[2]{\setlength{\tabcolsep}{#1}\renewcommand{\arraystretch}{#2}\centering\footnotesize}

\DeclareMathSymbol{@}{\mathord}{letters}{"3B}

\newcommand\mypara[1]{\vspace{0mm}\noindent\textbf{#1}}

\newcommand{\YesV}{\ding{51}}%
\newcommand{\NoX}{\ding{55}}%

\makeatletter
\DeclareRobustCommand\onedot{\futurelet\@let@token\@onedot}
\def\@onedot{\ifx\@let@token.\else.\null\fi\xspace}
\def\eg{\emph{e.g}\onedot} 
 
\def\cf{\emph{cf}\onedot} 
 \def\vs{\emph{vs}\onedot}

\newcommand*{\Rom}[1]{\expandafter\@slowromancap\romannumeral #1@}
\newcommand*{\rom}[1]{\expandafter\romannumeral #1}

\def\1{\bm{1}}

\def\rmA{{\mathbf{A}}}

\def\rmC{{\mathbf{C}}}

\def\rmF{{\mathbf{F}}}

\def\rmP{{\mathbf{P}}}

\def\vmu{{\bm{\mu}}}
\def\vnu{{\bm{\nu}}}

\def\vc{{\bm{c}}}

\def\vf{{\bm{f}}}
\def\vg{{\bm{g}}}

\def\vp{{\bm{p}}}
\def\vq{{\bm{q}}}

\def\vs{{\bm{s}}}

\def\vx{{\bm{x}}}

\def\mG{{\bm{G}}}

\def\mJ{{\bm{J}}}

\def\mP{{\bm{P}}}

\def\mR{{\bm{R}}}
\def\mS{{\bm{S}}}

\def\mW{{\bm{W}}}

\DeclareMathAlphabet{\mathsfit}{\encodingdefault}{\sfdefault}{m}{sl}
\SetMathAlphabet{\mathsfit}{bold}{\encodingdefault}{\sfdefault}{bx}{n}

\def\gC{{\mathcal{C}}}
\def\gD{{\mathcal{D}}}
\def\gE{{\mathcal{E}}}
\def\gF{{\mathcal{F}}}
\def\gG{{\mathcal{G}}}

\def\gI{{\mathcal{I}}}

\def\gL{{\mathcal{L}}}

\def\gN{{\mathcal{N}}}

\def\gP{{\mathcal{P}}}

\def\gS{{\mathcal{S}}}

\def\gV{{\mathcal{V}}}

\def\gX{{\mathcal{X}}}

\newcommand{\R}{\mathbb{R}}

\newcommand{\reg}{\lambda}

\let\originalleft\left
\let\originalright\right
\renewcommand{\left}{\mathopen{}\mathclose\bgroup\originalleft}
\renewcommand{\right}{\aftergroup\egroup\originalright}

\newcommand{\ours}{GaussianCross\xspace}
\newcommand{\scannet}{ScanNet\xspace}
\newcommand{\scannetp}{ScanNet200\xspace}
\newcommand{\structure}{Structure3D\xspace}
\newcommand{\sdis}{S3DIS\xspace}

\newcommand{\nerf}{NeRF\xspace}
\newcommand{\gs}{3DGS\xspace}

\AtBeginDocument{%
  }

\acmConference[MM '25]{Proceedings of the 33rd ACM International Conference on Multimedia}{October 27--31, 2025}{Dublin, Ireland}
\author{Lei Yao}
\email{rayyoh.yao@connect.polyu.hk}
\affiliation{
  \institution{Hong Kong Polytechnic University}
  \city{Hong Kong}
  \country{Hong Kong}
}

\author{Yi Wang}
\email{yi-eie.wang@polyu.edu.hk}
\authornotemark[1]
\affiliation{
  \institution{Hong Kong Polytechnic University}
  \city{Hong Kong}
  \country{Hong Kong}
}

\author{Yi Zhang}
\email{yi-eee.zhang@connect.polyu.hk}
\affiliation{
  \institution{Hong Kong Polytechnic University}
  \city{Hong Kong}
  \country{Hong Kong}
}

\author{Moyun Liu}
\email{lmomoy@hust.edu.cn}
\affiliation{
  \institution{Huazhong University of Science and Technology}
  \city{Wuhan}
  \country{China}
}

\author{Lap-Pui Chau}
\email{lap-pui.chau@polyu.edu.hk}
\authornote{Corresponding author}
\affiliation{
  \institution{Hong Kong Polytechnic University}
  \city{Hong Kong}
  \country{Hong Kong}
}

\begin{document}
\title{\ours: Cross-modal Self-supervised 3D Representation Learning via Gaussian Splatting}

\renewcommand{\shortauthors}{Yao et al.}
\renewcommand\footnotetextcopyrightpermission[1]{}

\begin{abstract}
  The significance of informative and robust point representations has been widely acknowledged for 3D scene understanding. Despite existing self-supervised pre-training counterparts demonstrating promising performance, the model collapse and structural information deficiency remain prevalent due to insufficient point discrimination difficulty, yielding unreliable expressions and suboptimal performance. In this paper, we present \textbf{\textit{\ours}}, a novel cross-modal self-supervised 3D representation learning architecture integrating feed-forward 3D Gaussian Splatting (\gs) techniques to address current challenges. \ours seamlessly converts scale-inconsistent 3D point clouds into a unified cuboid-normalized Gaussian representation without missing details, enabling stable and generalizable pre-training. Subsequently, a tri-attribute adaptive distillation splatting module is incorporated to construct a 3D feature field, facilitating synergetic feature capturing of appearance, geometry, and semantic cues to maintain cross-modal consistency. To validate \ours, we perform extensive evaluations on various benchmarks, including \scannet, \scannetp, and \sdis. In particular, \ours shows a prominent parameter and data efficiency, achieving superior performance through linear probing (<0.1\% parameters) and limited data training (1\% of scenes) compared to state-of-the-art methods. Furthermore, \ours demonstrates strong generalization capabilities, improving the full fine-tuning accuracy by 9.3\% mIoU and 6.1\% AP$_{50}$ on \scannetp semantic and instance segmentation tasks, respectively, supporting the effectiveness of our approach. The code, weights, and visualizations are publicly available at \href{https://rayyoh.github.io/GaussianCross/}{https://rayyoh.github.io/GaussianCross/}.
\end{abstract}

\begin{teaserfigure}
  \includegraphics[width=0.98\textwidth]{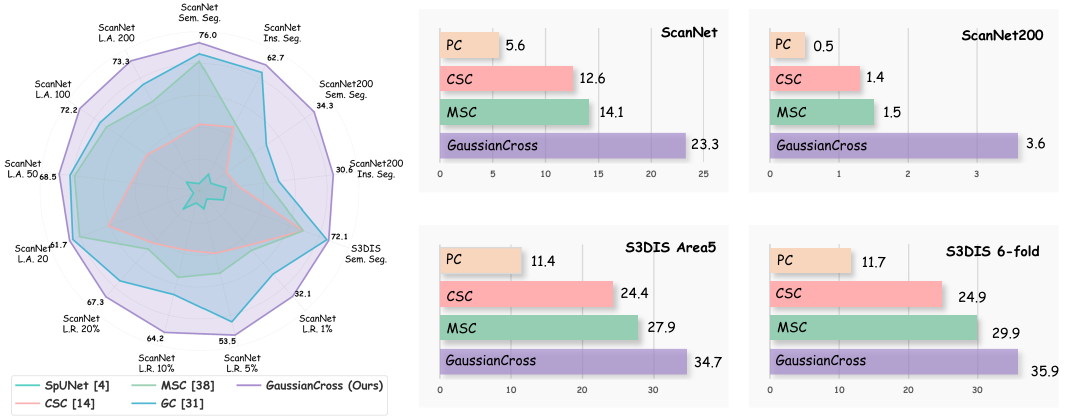}
  \caption{Performance comparison of \ours on 3D scene understanding tasks. \ours achieves superior performance across various tasks, including semantic segmentation (\textit{Sem. Seg.})~\cite{choy2019minkowski}, instance segmentation (\textit{Ins. Seg.})~\cite{jiang2020pointgroup}, and linear probing~\cite{wu2025sonata}. \textit{Left}: full fine-tuning results on various downstream tasks. \textit{Right}: linear probing accuracy.}
  \Description{Teaser figure.}
  \label{fig:teaser}
\end{teaserfigure}

\maketitle

\section{Introduction}
\label{sec:introduction}
Self-supervised representation learning has emerged as a transformative training paradigm for capturing expressive features from large-scale unlabeled data. It has demonstrated promising potential across diverse downstream applications, including scene understanding~\cite{yao2024sgiformer,liu2024menet}, navigation~\cite{zurbrugg2024icgnet}, and embodied manipulation~\cite{zheng2025survey}. While the success of 2D visual foundation models (VFMs) such as MAE~\cite{he2022masked}, MoCo~\cite{he2020momentum}, and DINOv2~\cite{oquab2023dinov2} trained by self-supervised pre-training, the development of comparable 3D methodologies remains critical for comprehensive physical world understanding~\cite{wu2025sonata}. However, different from available web-scale images, 3D data, especially point clouds, are usually scarce and come with sophisticated spatial structures, hindering the design of effective self-supervised representation learning strategies. The sparse and irregular nature of the point cloud further complicates the learning process.

Although recent investigations~\cite{pang2022masked, yu2022point, qi2023contrast, yu2024mm} have advanced object-level point cloud representation learning, these approaches face fundamental scale incompatibility when transitioning to scene-level scenarios. Concurrently, some frameworks~\cite{hou2021csc,xie2020pointcontrast,wu2023masked,wang2024groupcontrast,fan2024point} have attempted to explore contrastive learning-based algorithms for capturing compelling 3D scene features, which typically generate dual distinct views from the same scene and consider point-wise discrimination as their pretext tasks. Despite empirical improvements on downstream tasks, persistent challenges remain. For instance, PointContrast~\cite{xie2020pointcontrast} suffers from model collapse stemming from inadequate diversity in view augmentation strategies, while GroupContrast~\cite{wang2024groupcontrast} exhibits significant parameter sensitivity and depends on precomputed over-segmentations~\cite{felzenszwalb2004efficient}, thereby restricting its adaptability. On the other hand, the integration of neural rendering techniques introduces alternative pathways for self-supervised representation learning. Ponder~\cite{huang2023ponder} pioneers a Neural Radiance Field (\nerf)~\cite{mildenhall2021nerf} based pre-training paradigm that leverages novel view synthesis as the supervisory signal, but its practical scalability is hampered by the inherent slow training and rendering speed. GS$^3$~\cite{liu2024point} conducts a preliminary exploration of 3D Gaussian Splatting~\cite{kerbl20233d} (\gs) for rendering-based pre-training strategy, which implements epipolar transformer~\cite{wang2024freesplat} for cross-view pixel-wise alignment. However, this approach focuses exclusively on photometric reconstruction while neglecting critical geometric and semantic relationships, resulting in suboptimal performance on structurally complex downstream tasks. Additionally, the method starts from back-projected point clouds of sparse view RGB-D frames, which is inherently limited to global context modeling.

To address the aforementioned challenges, we propose \textit{\textbf{\ours}}, a novel cross-modal self-supervised 3D representation learning framework with Gaussian Splatting to learn informative and robust point representations for scene understanding. Unlike the per-scene optimization paradigm of vanilla \gs~\cite{kerbl20233d}, our method operates in a generalizable manner and is tailored to capture diverse intrinsic properties. Nevertheless, a potential challenge is scale uncertainty across different indoor scenes, which causes the model struggling to learn a unified representation as shown in~\figref{fig:ablation} top (\textit{w/o} Cuboid-Normalized). To this end, we propose \textit{Cuboid-Normalized Gaussian Initialization}, a technique leveraged to transform scene point clouds into a cuboid structure and parameterize them as a collection of Gaussian primitives. The process enables the model to flexibly adapt to scale variations in different scenes, allowing seamless scene description conversion without compromising detail fidelity. Furthermore, we introduce a \textit{Tri-Attribute Adaptive Distillation Splatting} module that utilizes the real-time rendering capability of rasterization splatting~\cite{kerbl20233d}. Apart from common Gaussian characteristics, we predict an offset to dynamically refine the mean position and integrate an opacity-driven pruning mechanism to control primitive density, which has proved crucial for accurate scene representation. In addition, we incorporate a 3D feature field to guide semantic map synthesis, aiming to pursue high-level semantic-aware details. The generated maps are then upsampled by a projection head to align with latent embeddings of a pre-trained 2D foundation model, facilitating cross-modal knowledge distillation. \ours achieves simultaneous capture of complementary photometric appearance, geometric structure, and semantic context, prompting synergistic feature learning. The self-supervised training process is performed by reconstructing randomly sampled views to provide robust supervision, effectively mitigating model collapse risk. Our contributions comprise:
\begin{itemize}
    \item We propose a novel cross-modal self-supervised 3D representation learning architecture for scene understanding with generalizable Gaussian Splatting, named \ours.
    \item We introduce a cuboid-normalized Gaussian initialization technique to represent scenes as structured 3D Gaussians, adapting to inconsistent scales across different scenes.
    \item We design a tri-attribute adaptive distillation splatting module to jointly capture the appearance, geometry, and semantic properties of scenes, achieving cross-modal knowledge distillation from visual foundation models.
    \item Comprehensive experiments on various scene understanding tasks demonstrate the superior performance of \ours over previous state-of-the-art methods.
\end{itemize}

\section{Related work}
\label{sec:related}

\subsection{Point Clouds Self-supervised Learning}
The recent proliferation of self-supervised learning in 2D~\cite{heinrich2025radiov25,zhu2025emosym} has inspired research efforts to adapt this paradigm to point cloud analysis. Pioneering works like Point-MAE~\cite{pang2022masked} and Point-BERT~\cite{yu2022point} successfully transferred masked autoencoding~\cite{devlin2019bert} to object-level point clouds by transformer-based  architectures~\cite{vaswani2017attention}. However, scaling such object-centric approaches to scene tasks is non-trivial due to sparse geometric structures in real-world 3D scenes. To address this challenge, PointContrast (PC)~\cite{xie2020pointcontrast} established an unsupervised framework for indoor scenes, which learns point-wise representation derived from RGB-D frames by maximizing the mutual information between augmented views. Building upon this foundation, Contrastive Scene Context (CSC)~\cite{hou2021csc} introduced spatial contextual constraints to encode structural relationships beyond individual points correspondence. In~\cite{wu2023masked}, Masked Scene Contrast (MSC) unified color reconstruction and surfel normal prediction within a pipeline and proposed an efficient view generation strategy. In contrast, recent innovations highlight semantic-aware learning as a critical frontier. For example, GroupContrast (GC)~\cite{wang2024groupcontrast} identified the semantic ambiguity problem and addressed it by a segment grouping strategy based on pre-computed superpoints~\cite{felzenszwalb2004efficient}. It further proposed a group-aware contrastive loss to enhance the representation, while Point-GCC~\cite{fan2024point} incorporated deep clustering for object-level supervision. Despite these advancements, current contrastive methods remain susceptible to model collapse phenomena~\cite{wang2024groupcontrast} and exhibit parametric sensitivity. Our approach diverges from them by leveraging a cross-modal pre-training paradigm, which enhances robustness and generalizability.

\subsection{Cross-modal 3D Pre-training}
There is another series of works aiming to pre-train 3D models with cross-modal data. MM-Point~\cite{yu2024mm} enforced cross-modal consistency representations through point-to-pixel projection, aligning specific view images with point clouds. While effective, these methods critically rely on the availability of well-aligned 2D-3D pairs, which may not be feasible in many real-world applications. Instead, some recent works~\cite{huang2023ponder, zhu2023ponderv2, liu2024point} consider differentiable rendering as a self-supervised signal by comparing arbitrary synthetic views with real images for 3D scenes. Ponder~\cite{huang2023ponder} employed the neural radiance fields-based~\cite{mildenhall2021nerf} technique for SDF values and colors prediction from query points based on NeuS~\cite{wang2021neus}. Subsequent work GS$^3$~\cite{liu2024point} adopted 3D Gaussian Splatting~\cite{kerbl20233d} for photorealistic rendering starting from multi-view RGB-D frames, but this approach required input views to have overlapped regions and additional computational cost due to its epipolar transformer~\cite{wang2024freesplat} for view alignment. PonderV2~\cite{zhu2023ponderv2} extended the prior version~\cite{huang2023ponder} to multi-source pre-training based on Point Prompt Training (PPT)~\cite{wu2024towards} with language-guided alignment. Nevertheless, a potential limitation is its reliance on 2D ground-truth supervision, which hinders its scalability. Our work establishes another paradigm in this domain through semantic-aware knowledge distillation from VFMs to point clouds with feed-forward Gaussian splatting, enabling effective pre-training without any annotations.


\subsection{Generalizable 3D Gaussian Splatting}
Neural Radiance Fields (NeRF)~\cite{mildenhall2021nerf} implicitly represent 3D scenes with shallow Multi-Layer Perceptrons (MLPs), learning continuous mappings from spatial coordinates to radiance fields. However, the necessity of dense point sampling imposes a significant computational burden during both the training and rendering phases. 3D Gaussian Splatting (3DGS)~\cite{kerbl20233d} revolutionized this paradigm by explicit scene parameterization using anisotropic Gaussian primitives, achieving real-time rendering via differentiable rasterization splatting. Although its high-quality rendering output, 3DGS is limited to scene-specific optimization and lacks the ability to generalize to unseen scenes~\cite{charatan2024pixelsplat}. To address this problem, anchor-based 3DGS methods~\cite{charatan2024pixelsplat, chen2024mvsplat, wang2024freesplat} are proposed. Specifically, PixelSplat~\cite{charatan2024pixelsplat} incorporated epipolar transformers into the pipeline to enable a feed-forward training paradigm for generalizable 3DGS, while MVSplat~\cite{chen2024mvsplat} and FreeSplat~\cite{wang2024freesplat} introduced additional techniques to construct cost volume for efficient training and free-viewpoint rendering. Parallel advancements focus on enhancing Gaussian representations through cross-modal fusion. GaussianGrouping~\cite{gaussian_grouping} integrated priors for part-aware decomposition, Feature-3DGS~\cite{zhou2024feature} established dense 2D-3D feature correspondences, and FiT3D~\cite{yue2024improving} adapted visual foundation models via 3D-aware fine-tuning. Inspired by these works, our \ours introduces a novel knowledge distillation framework that transfers VFM-derived semantic features into geometrically grounded Gaussian embeddings, enabling label-efficient pre-training of point cloud encoders.

\section{Methodology}
\label{sec:method}
\begin{figure*}
    \centering
    \includegraphics[width=0.99\linewidth]{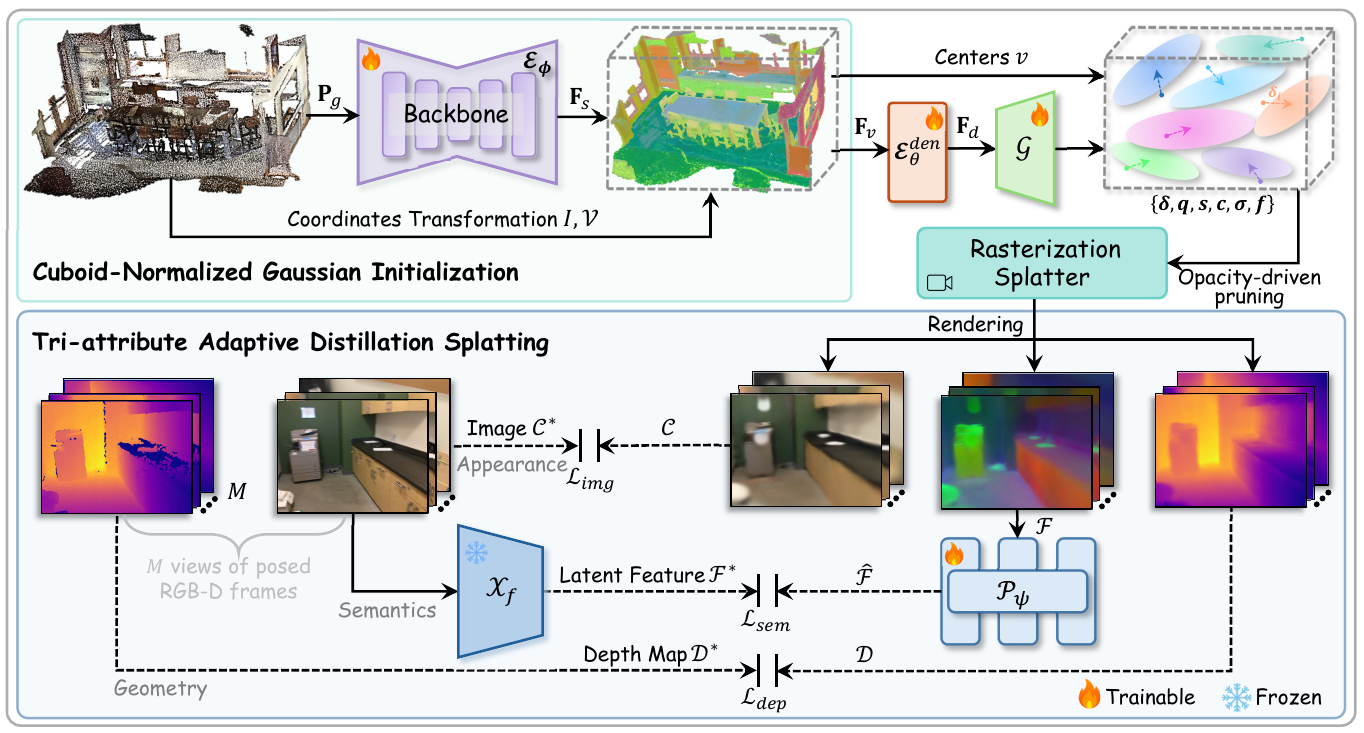}
    \caption{The overall architecture of \ours. The pipeline commences with cuboid-normalized Gaussian initialization to establish coarse primitive means. Gaussian properties are subsequently decoded by $\gG$ with a feature field. The tri-attribute adaptive distillation splatting is performed to ensure cross-modal consistency.}
  \Description{Architecture.}
    \label{fig:architecture}
\end{figure*}

This section begins with the preliminaries of \gs and presents the overall architecture of \ours in~\figref{fig:architecture}. We subsequently detail our cuboid-normalized Gaussian initialization in~\secref{subsec:param} and introduce the tri-attribute adaptive distillation splatting in~\secref{subsec:rendering}. Finally, we describe the loss functions in~\secref{subsec:loss} that regularize our cross-modal self-supervised learning.

\subsection{Preliminaries}
\label{subsec:preliminaries}
\gs~\cite{kerbl20233d} considers a cluster of translucent ellipsoids characterized by Gaussian primitives to represent scenes explicitly. Each of them is defined by a center $\vmu \in \R^3$ and covariance matrix $\bm{\Sigma} \in \R^{3 \times 3}$, expressed as:
\begin{equation}
    \mG(\vx) = e^{-\frac{1}{2}(\vx - \vmu)^T\bm{\Sigma}^{-1}(\vx - \vmu)}.
    \label{eq:gaussian}
\end{equation}
To assure positive semi-definiteness during differentiable optimization, $\bm{\Sigma}$ is decomposed as $\bm{\Sigma} = \mR\mS\mS^T\mR^T$, where $\mR = \texttt{q2r}(\vq)$ and $\mS = \texttt{diag}(\vs)$ are rotation and scaling matrices, respectively. The operators $\texttt{q2r}(\cdot)$ and $\texttt{diag}(\cdot)$ convert quaternions to rotation matrices and construct diagonal matrices from scaling vectors, respectively. Given an arbitrary view transformation matrix $\mW$, the 3D Gaussians are splatted onto specific 2D camera plane with corresponding mean and covariance:
\begin{equation}
    \vmu_{2D} = \mP\mW\vmu, \quad \bm{\Sigma}_{2D} = \mJ\mW\bm{\Sigma}\mW^T\mJ^T,
    \label{eq:2d_cov}
\end{equation}
where $\mP$ denotes projective transformation and $\mJ$ the Jacobian. Final pixel color is computed by alpha-blending $\gN$ ordered Gaussians:
\begin{equation}
    \bm{\gC}(\vp) = \sum_{i \in \gN} \vc_i \alpha_i \prod_{j = 1}^{i - 1} (1 - \alpha_j),
    \label{eq:color}
\end{equation}
where $\vc_i$ represents view-dependent spherical harmonics color and $\alpha_i$ combines $\bm{\Sigma}_{2D}$ with opacity $\bm{\sigma}_i$. 

\subsection{Cuboid-Normalized Gaussian Initialization}
\label{subsec:param}
This section investigates the integration of \gs into point cloud representation learning, motivated by its promise in complex scene modeling without requiring labor-intensive 3D annotations. However, conventional \gs methods face limitations in scale-variant scenes representation due to their scene-specific optimization. Inspired by~\cite{zhu2023ponderv2}, we propose cuboid-normalized Gaussian initialization aiming to alleviate scale variance effects while enabling generalizable feature learning directly from point input.

Given a raw scene point cloud $\rmP_r = \{\rmC_{r,i}, \rmA_{r,i}\}_{i=1}^n$, where $\rmC_{r,i} \in \R^3$ denotes spatial coordinates $x_i, y_i, z_i$ and $\rmA_{r,i} \in \R^c$ represents associated $c$-dimensional attributes (\eg RGB colors, surface normals) per point. Analogous to previous works~\cite{wu2023masked,liu2024point}, we mask out a portion of the input by a ratio $\gamma$ and apply a sampling pattern:
\begin{equation}
    \gS: \left\lfloor \gamma \rmP_r \right\rfloor \mapsto \rmP_g = \{\rmC_{g,i}, \rmA_{g,i}\}_{i=1}^m
\end{equation}
with the size $\vg$ to downsample the point cloud from $n$ to $m$ points. The subsampled point cloud $\rmP_g$ is subsequently processed by a 3D backbone $\gE_{\bm{\phi}}$ with learnable parameters $\bm{\phi}$: 
\begin{equation}
    \rmF_{s} = \gE_{\bm{\phi}}(\rmP_g) \in \R^{m \times d_s},
\end{equation}
yielding sparse features where $d_s$ is the channel dimension. Our objective centers on learning discriminative and reliable point-wise representations through $\gE_{\bm{\phi}}$ by leveraging cross-modal self-supervision signals.

To construct scale-agnostic representations, we develop a normalized cuboid volumetric encoding scheme. This spatial normalization is essential for learning generalizable scene representations across varying scales. Specifically, we perform coordinate transformation $\gI$ to map raw positions $\rmC_g$ into a unit cube, which guarantees all scenes occupy a canonical domain while preserving relative spatial relationships. We further apply a discretization operation $\gV$, partitioning the cube into $X \times Y \times Z$ uniformly voxels. This process is described in~\eqref{eq:voxelization} and voxel centers $\rmC_v$ are given by:
\begin{equation}
    \rmC_v = \gV \left ( \gI(\rmC_g), X, Y, Z \right ).
    \label{eq:voxelization}
\end{equation}
Each point $\rmC_{g,i}$ is assigned a unique voxel index $id \in \{1, 2, \ldots, X \times Y \times Z\}$ determined by spatial hashing and grid resolution, yielding an index set $ids = \{id_i\}_{i=1}^m$. The voxel-wise embeddings are then attained by scattering sparse features sharing identical indices:
\begin{equation}
    \rmF_v = \texttt{Scatter} \left (\rmF_s, \gI(\rmC_g), ids, \rmC_v \right )\in \R^{X \times Y \times Z \times d_s}
    \label{eq:voxel_scattering}
\end{equation}
where unoccupied voxels are filled with zeros. The features $\rmF_v$ are then processed by a 3D convolutional neural network $\gE_{\bm{\theta}}^{den}$ to establish a dense feature volume:
\begin{equation}
    \rmF_d = \gE_{\bm{\theta}}^{den}(\rmF_v) \in \R^{X \times Y \times Z \times d_o},
\end{equation}
where $d_o$ denotes the output dimension. With the structured scene representation, we consider each voxel as an anchor and directly serve its center $\rmC_{v,i}$ as coarse mean $\vnu_i$ of the Gaussian. The voxel features $\rmF_{d,i}$ are also assigned to the $i$-th Gaussian.  Our experiments demonstrate this cuboid-normalized initialization empirically outperforms traditional SfM-based \gs methods~\cite{kerbl20233d,snavely2006photo} in representation consistency (see~\figref{fig:ablation}), effectively enabling direct Gaussian initialization from raw point clouds.

\subsection{Tri-attribute Adaptive Distillation Splatting}
\label{subsec:rendering}
To achieve self-supervised 3D representation learning, we consider novel view synthesis as a pretext task, eliminating dependency on 3D supervision while maximally utilizing available 2D data. Building upon the dense features $\rmF_d$ obtained in~\secref{subsec:param}, we parameterize Gaussian attributes via dedicated Multi-Layer Perceptrons (MLPs) decoders with associated activations:
\begin{equation}
    \vq_i = {Normalize}\left(\gG_q(\rmF_{d,i})\right), \quad \vs_i = Softplus\left(\gG_s(\rmF_{d,i})\right),
    \label{eq:quaternion_scaling}
\end{equation}
where $\gG_q$ and $\gG_s$ are quaternion and scaling prediction heads. Color $\vc_i$ and opacity $\bm{\sigma}_i$ are similarly decoded by:
\begin{equation}
    \vc_i = Sigmoid\left(\gG_c(\rmF_{d,i})\right), \quad \bm{\sigma}_i = Sigmoid\left(\gG_{\sigma}(\rmF_{d,i})\right).
    \label{eq:color_opacity}
\end{equation}
To address inaccuracy of coarse mean $\vnu_i$ initialization in representing the actual scene, we introduce a predicted offset $\bm{\delta}_i$ by:
\begin{equation}
    \bm{\delta}_i = tanh\left(\gG_{\delta}(\rmF_{d,i})\right) \cdot \Delta.
    \label{eq:offset}
\end{equation}
Here, $\Delta$ controls the maximum displacement magnitude. The learned offset $\bm{\delta}_i$ is then added to $\vnu_i$ yielding the refined mean $\vmu_i = \vnu_i + \bm{\delta}_i$. Concurrently, we establish a feature field to capture potential semantic cues of each anchor by projecting the dense features $\rmF_{d,i}$ into a semantic-aware embedding $\vq_i$ with the dimension of $d_q$:
\begin{equation}
    \vf_i = \gG_f(\rmF_{d,i}),
    \label{eq:semantic_prediction}
\end{equation}
These attributes enable modeling the scene from different perspectives and capturing comprehensive information. Although directly initializing Gaussians from voxels ensures training efficiency, inherent redundancy may compromise rendering fidelity and computational efficiency. We therefore introduce an opacity-driven pruning mechanism with a threshold $\tau$ to determine whether reserving the anchor. Finally, we can explicitly represent the 3D scene by a series of Gaussian primitives characterized by predicted properties:
\begin{equation}
    \left\{\vmu_i, \vq_i, \vs_i, \vc_i, \bm{\sigma}_i, \vf_i \mid \bm{\sigma}_i > \tau \right\}_{i=1}^{X \times Y \times Z}.
\end{equation}

Then, we propose tri-attribute adaptive distillation splatting to render multi-view images, depth, and feature maps, enabling the model to pursue underlying photometric appearance, geometric structure, and semantic information. The splatting is performed by projecting 3D Gaussian primitives onto $M$ camera planes with different poses. Instead of picking specific views like~\cite{wang2024pfgs}, we randomly sample $M$ views from the training dataset for each scene to enhance generalization ability. Color outputs $\{\bm{\gC}_m\}_{m=1}^M$ are synthesized following \eqref{eq:color}, where $\bm{\gC}_m \in \R^{H \times W \times 3}$, $H$ and $W$ are height and width. Subsequently, geometric regularization is established by depth map $\bm{\gD}_m \in \R^{H \times W}$ generation:
\begin{equation}
    \bm{\gD}_m(\vp) = \sum_{i \in \gN} d_i \alpha_i \prod_{j = 1}^{i - 1} (1 - \alpha_j),
\end{equation}
where $d_i$ is the camera space $z$-depth of the $i$-th Gaussian. Our framework further integrates feature field rendering into the procedure to distill semantic-aware knowledge from a 2D visual foundation model. Unlike PonderV2~\cite{zhu2023ponderv2} that directly predicts 2D semantic labels, we consider feature correlations as intermediate supervision to guide feature learning, eliminating the requirement of ground-truth labels. The rendered feature map $\bm{\gF}_m \in \R^{H \times W \times d_f}$ is denoted as:
\begin{equation}
    \bm{\gF}_m(\vp) = \sum_{i \in \gN} \vf_i \alpha_i \prod_{j = 1}^{i - 1} (1 - \alpha_j).
\end{equation}
We employ the latent features from a pre-trained VFM $\gX_{f}$ as the prior: $\bm{\gF}^{*}_m = \gX_{f}(\bm{\gC}^{*}_m) \in \R^{H \times W \times d^{*}}$,
where $\gX_{f}$ is an arbitrary 2D foundation model and $\bm{\gC}^{*}_m$ is the corresponding real color image. Nevertheless, a potential challenge lies in that the dimension $d^{*}$ of $\bm{\gF}^{*}_m$ is usually large, making it time-consuming to render such high-dimensional feature maps. Therefore, we tend to render a low-dimensional map ($d_f \ll d^*$). To address the dimension disparity, we implement a lightweight projection head $\gG_{\bm{\psi}}$ to upsample $\bm{\gF}_m$ to align with the dimension of $\bm{\gF}^{*}_m$:
\begin{equation}
    \bm{\hat{\gF}}_m = \gP_{\bm{\psi}}(\bm{\gF}_m),
\end{equation}
where $\bm{\hat{\gF}}_m \in \R^{H \times W \times d^{*}}$. This design strategically balances computational efficiency with semantic fidelity, enabling effective distillation of 2D priors into 3D representations without compromising rendering performance.

\subsection{Training Loss Functions}
\label{subsec:loss}
The principle of our design is to adhere the model to capture multifaceted properties from raw 3D scenes and incorporate available priors from VFMs into 3D feature space. We introduce a $l_1$ loss denoted as $\gL_{img}$ to measure the discrepancy of exported photorealistic images $\bm{\gC}_m$ and the ground truth $\bm{\gC}_{m}^{*}$ aiming to capture adequate appearance details:
\begin{equation}
    \gL_{img} = \frac{1}{M} \sum_{m=1}^M \lVert \bm{\gC}_m - \bm{\gC}_{m}^{*} \rVert.
\end{equation}
For splatted depth maps $\bm{\gD}_m$, we also use the $l_1$ loss $\gL_{dep}$ within valid pixels to regularize geometric features alignment with concomitant real depth maps $\bm{\gD}_{m}^{*}$:
\begin{equation}
    \gL_{dep} = \frac{1}{M \cdot H W} \sum_{m=1}^M \sum_{h=1}^{H} \sum_{w=1}^{W} \mathbb{I}_{\{\bm{\gD}_{m,h,w}^{*}\}} \lVert \bm{\gD}_{m,h,w} - \bm{\gD}_{m,h,w}^{*} \rVert.
\end{equation}
where $\mathbb{I}_{\{\cdot\}}$ denotes the indicator function. Furthermore, in terms of the yielded feature maps $\bm{\hat{\gF}}_m$ from our semantic feature field, we integrate a similarity loss $\gL_{sem}$ to distill 2D knowledge priors by aligning with $\bm{\gF}^{*}_m$ from VFMs:
\begin{equation}
    \gL_{sem} = \frac{1}{M} \sum_{m=1}^M \left[1 - \frac{\bm{\hat{\gF}}_m \cdot \bm{\gF}^{*}_m}{\lVert \bm{\hat{\gF}}_m \rVert \lVert \bm{\gF}^{*}_m \rVert} \right].
\end{equation}
Therefore, our cross-modal pre-training framework can work in a self-supervised manner without the requirement of human annotations, and the total loss is defined as:
\begin{equation}
    \gL = \reg_{img} \gL_{img} + \reg_{dep} \gL_{dep} + \reg_{sem} \gL_{sem},
\end{equation}
where $\reg_{img}$, $\reg_{dep}$, and $\reg_{sem}$ are weights to balance different losses. 

\section{Experiments}
\label{sec:experiment}

\subsection{Experimental Settings}

\mypara{Backbone and Data.} 
We implement our \ours by Pointcept~\cite{pointcept2023}. Following established practice~\cite{wang2024groupcontrast, zhu2023ponderv2}, we adopt a Submanifold Sparse Convolution UNet~\cite{graham2017submanifold} (SparseUNet) as the 3D backbone $\gE_{\bm{\phi}}$ and consider 6-dimensional attributes as input features, comprising RGB values and normal vectors. We pre-train \ours on \scannet~\cite{dai2017scannet} and evaluate downstream scene understanding performance on \scannet, \scannetp~\cite{rozenberszki2022lground}, and \sdis~\cite{armeni2016s3dis} benchmarks, respectively. \textit{\scannet}~\cite{dai2017scannet} provides 1601 3D scenes with corresponding RGB-D frames, including 20 semantic classes for semantic segmentation and 18 object categories for instance recognition. The extended challenging version, \textit{\scannetp}~\cite{rozenberszki2022lground}, shares the same data yet contains more fine-grained annotations, expanding the labels to 200 semantic categories and 198 instance types. \textit{\sdis} complements our evaluation with 271 indoor scans across 6 large-scale areas, annotated with 13 distinct classes. We evaluate the performance on Area5 and 6-fold cross-validation settings.

\mypara{Training Details.}
We train \ours on \scannet~\cite{dai2017scannet} for 1200 epochs using 8 NVIDIA RTX 4090 GPUs with a batch size of 32. The learning rate is initialized as $2e^{-3}$ with the AdamW optimizer, modulated by a OneCycle learning rate scheduling policy. Input point clouds undergo standard geometric augmentations, including random rotation, anisotropic scaling, and flipping. Our view synthesis configuration uses 5 rendering views, each with a resolution of 480 $\times$ 640. The mask ratio $\gamma$ is set to 50\%, and the opacity threshold $\tau$ is set to 0.3 to trade-off between rendering fidelity and computational efficiency. For semantic feature alignment, we integrate pre-trained weight from RADIOv2.5~\cite{heinrich2025radiov25} as the frozen visual encoder $\gX_{f}$.

\subsection{Comparison with State-of-the-Art Methods}
In this section, we conduct comprehensive benchmarking of \ours against existing approaches across various tasks. We start by assessing parameter efficiency by linear probing following the protocol established in Sonata~\cite{wu2025sonata} and data efficiency with limited scene reconstruction and point annotation data settings. We then evaluate the transfer learning performance through full fine-tuning on 3D semantic and instance segmentation tasks. In our tables, we denote \scratch, \pretrain, and \suppretrain as training from scratch, self-supervised pre-training, and supervised pre-training, respectively. For more details, please refer to the supplementary materials.

\begin{table}[t]
    \centering
    \caption{Parameter efficiency via linear probing. \textit{SpUNet} means SparseUNet~\cite{graham2017submanifold} as the backbone.}
    \tablestyle{1.5pt}{1.0}
    \resizebox{0.48\textwidth}{!}{
    \begin{tabular}{y{14mm}x{5mm}x{5mm}x{5mm}x{6mm}x{5mm}x{6mm}x{6mm}x{6mm}x{5mm}}
        \toprule
        \textbf{\textit{Linear Prob.}}& \multicolumn{2}{c}{\textit{ScanNet}} & \multicolumn{2}{c}{\textit{ScanNet200}} & \multicolumn{2}{c}{\textit{S3DIS Area5}} & \multicolumn{2}{c}{\textit{S3DIS 6-fold}} \\
        \cmidrule(lr){1-1} \cmidrule(lr){2-3} \cmidrule(lr){4-5} \cmidrule(lr){6-7} \cmidrule(lr){8-9}
        Methods & mIoU & mAcc & mIoU & mAcc & mIoU & mAcc & mIoU & mAcc \\
        \midrule
        \rowcolor{gray} \scratch SpUNet~\cite{choy2019minkowski} & 72.2 & 80.2 & 25.0 & 32.9  & 66.3 & 72.5 & 72.4 & 80.9 \\
        \midrule
        \pretrain PC~\cite{xie2020pointcontrast} & 5.6 & 9.7 & 0.5 & 0.9 & 11.4 & 18.6 & 11.7 & 19.0 \\
        \pretrain CSC~\cite{hou2021csc} & 12.6 & 18.1 & 1.3 & 2.1 & 24.4 & 32.0 & 24.9 & 32.5 \\
        \pretrain MSC~\cite{wu2023masked} & 14.1 & 20.3 & 1.5 & 2.5 & 27.9 & 35.5 & 29.9 & 37.9 \\
        \midrule
        \rowcolor{ourscolor} \pretrain \textbf{Ours} & \textbf{23.3} & \textbf{30.9} & \textbf{3.6} & \textbf{5.3} & \textbf{34.7} & \textbf{44.1} & \textbf{35.9} & \textbf{45.5} \\
        \bottomrule
    \end{tabular}
}
    \label{tab:sem_seg_lin}
\end{table}

\subsubsection{\textbf{Linear Probing}}
To quantify the intrinsic quality of learned representations, we implement a linear evaluation protocol where only the classification layer undergoes training while the backbone remains frozen. This parameter-efficient paradigm directly measures feature separability in the pre-trained embedding space. Results in~\tabref{tab:sem_seg_lin} demonstrate \ours's superiority, achieving 23.3\%, 3.6\%, 34.7\%, 35.9\% mIoU on \scannet, \scannetp, \sdis Area5 and 6-fold, respectively. Although \ours outperforms other methods, the performance discrepancy between linear probing and full training reveals that current self-supervised objectives remain to be further optimized. This suggests that while \ours excels in learning transferable representations, there is still room for improvement in the pre-training process itself.

\begin{table}[t]
    \centering
    \caption{Data efficiency on \scannet Data Efficient benchmark~\cite{hou2021csc} by limited scenes and point annotations.}
    \tablestyle{1.5pt}{1.0}

\resizebox{0.48\textwidth}{!}{
    \begin{tabular}{y{14mm}x{5mm}x{5mm}x{5mm}x{6mm}x{6mm}x{5mm}x{5mm}x{6mm}}
        \toprule
        \textbf{\textit{Data Eff.}} & \multicolumn{4}{c}{\textit{Limited Scenes (Pct.)}} & \multicolumn{4}{c}{\textit{Limited Annotations (Pts.)}} \\
        \cmidrule(lr){1-1} \cmidrule(lr){2-5} \cmidrule(lr){6-9}
        Methods & 1\% & 5\% & 10\% & 20\% & 20 & 50 & 100 & 200 \\
        \midrule
        \rowcolor{gray} \scratch SpUNet~\cite{choy2019minkowski} & 26.0 & 47.8 & 56.7 & 62.9 & 41.9 & 53.9 & 62.2 & 65.5 \\
        \midrule
        \pretrain CSC~\cite{hou2021csc} & 28.9 & 49.8 & 59.4 & 64.6 & 55.5 & 60.5 & 65.9 & 68.2 \\
        \pretrain MSC~\cite{wu2023masked} & 29.2 & 50.7 & 61.0 & 64.9 & 60.1 & 66.8 & 69.7 & 70.7 \\
        \pretrain GC~\cite{wang2024groupcontrast} & 30.7 & \underline{52.9} & 62.0 & \underline{66.5} & \underline{61.2} & 67.3 & 70.3 & 71.8 \\
        \suppretrain PPT~\cite{wu2024towards} & \underline{31.3} & 52.3 & \underline{62.8} & 66.4 & 60.6 & \underline{67.5} & \underline{70.8} & \underline{72.2} \\
        \midrule
        \rowcolor{ourscolor} \pretrain \textbf{Ours} & \textbf{32.1} & \textbf{53.5} & \textbf{64.2} & \textbf{67.3} & \textbf{61.7} & \textbf{68.5} & \textbf{72.2} & \textbf{73.3} \\
        \hspace{0.8em} $\Delta$ & \textcolor{offsetcolor}{\textbf{+6.1}} & \textcolor{offsetcolor}{\textbf{+5.7}} & \textcolor{offsetcolor}{\textbf{+7.5}} & \textcolor{offsetcolor}{\textbf{+4.4}} & \textcolor{offsetcolor}{\textbf{+19.8}} & \textcolor{offsetcolor}{\textbf{+14.6}} & \textcolor{offsetcolor}{\textbf{+10.0}} & \textcolor{offsetcolor}{\textbf{+7.8}} \\
        \bottomrule
    \end{tabular}
}
    \label{tab:sem_seg_data}
\end{table}

\subsubsection{\textbf{Data Efficiency}}
In~\tabref{tab:sem_seg_data}, we systematically evaluate the data efficiency by fine-tuning on \scannet Data Efficient benchmark~\cite{hou2021csc} with limited scenes and point annotations. The results on both configurations exhibit impressive improvements compared to learning from scratch baselines (\cf \scratch). In the case of extreme data scarcity and limited point annotations, \ours also obtains the best performance among all other counterparts, with 32.1\% and 61.7\% mIoU on 1\% scenes and 20 points per scene scenarios. Notably, \ours can even outperform the supervised pre-training model (\eg \suppretrain PPT~\cite{wu2024towards}), providing empirical validation that our cross-modal self-supervised objectives learn more transferable structural priors than manually curated supervision. This evidence positions \ours as a theoretically grounded framework for label-efficient 3D scene understanding.

\begin{table*}[!htbp]
    \centering
    \caption{3D semantic segmentation results. The best results are highlighted in \textbf{bold}, and the second-best results are in \underline{underlined}.}
    \tablestyle{1.5pt}{1.0}
    \resizebox{0.99\textwidth}{!}{
    \begin{tabular}{y{19mm}|x{15mm}|x{30mm}|x{12mm}|x{10mm}x{11mm}x{8mm}x{8mm}}
        \toprule
        \multicolumn{4}{c}{\textbf{\textit{Semantic Segmentation}}} & \scannet & \scannetp & \multicolumn{2}{c}{\sdis} \\
        \cmidrule(lr){1-4} \cmidrule(lr){5-5} \cmidrule(lr){6-6} \cmidrule(lr){7-8} 
        Methods & Venue & Pre-training Datasets & Type & Val mIoU & Val mIoU & Area5 & 6-fold \\
        \midrule
        \multicolumn{8}{c}{\textit{Supervised Learning from Scratch}} \\
        \midrule
        \scratch PointNeXt~\cite{qian2022pointnext} & NeurIPS 2022 & \NoX & \NoX & 71.5 & - & 70.5 & 74.9 \\
        \scratch StFormer~\cite{lai2022stratified} & CVPR 2022 & \NoX & \NoX & 74.3 & - & 72.0 & - \\
        \scratch PTv1~\cite{zhao2021pointtransformer} & ICCV 2021 & \NoX & \NoX & 70.6 & 27.8 & 70.4 & 65.4 \\
        \scratch PTv2~\cite{wu2022pointtransformerv2} & NeurIPS 2022 & \NoX & \NoX & 75.4 & 30.2 & 71.6 & 75.1 \\
        \rowcolor{gray} \scratch SpUNet~\cite{choy2019minkowski} & CVPR 2019 & \NoX & \NoX & {72.2} & {25.0} & {66.3} & 72.4 \\
        \midrule
        \multicolumn{8}{c}{\textit{Self-supervised Pre-training}} \\
        \midrule
        \pretrain GS$^3$~\cite{liu2024point} & arXiv 2024 & \scannet & Rendering & 73.4$_{\textcolor{darkgray}{\textbf{+1.2}}}$ & - & 70.1$_{\textcolor{darkgray}{\textbf{+3.8}}}$ & - \\
        \pretrain Ponder~\cite{huang2023ponder} & CVPR 2023 & \scannet & Rendering & 73.5$_{\textcolor{darkgray}{\textbf{+1.3}}}$ & - & - & - \\
        \pretrain CSC~\cite{hou2021csc} & CVPR 2021 & \scannet & Contrast & 73.8$_{\textcolor{darkgray}{\textbf{+1.6}}}$ & 26.4$_{\textcolor{darkgray}{\textbf{+1.4}}}$ & 70.7$_{\textcolor{darkgray}{\textbf{+4.4}}}$ & \underline{75.5}$_{\textcolor{darkgray}{\textbf{+3.1}}}$ \\
        \pretrain PC~\cite{xie2020pointcontrast} & ECCV 2020 & \scannet & Contrast & 74.1$_{\textcolor{darkgray}{\textbf{+1.9}}}$ & 26.2$_{\textcolor{darkgray}{\textbf{+1.2}}}$ & 70.3$_{\textcolor{darkgray}{\textbf{+4.0}}}$ & 74.7$_{\textcolor{darkgray}{\textbf{+2.3}}}$ \\
        \pretrain MSC~\cite{wu2023masked} & CVPR 2023 & \scannet, ArkitScenes & Contrast & 75.5$_{\textcolor{darkgray}{\textbf{+3.3}}}$ & \underline{32.0}$_{\textcolor{darkgray}{\textbf{+7.0}}}$ & 70.7$_{\textcolor{darkgray}{\textbf{+4.4}}}$ & - \\
        \pretrain GC~\cite{wang2024groupcontrast} & CVPR 2024 & \scannet & Contrast & 75.7$_{\textcolor{darkgray}{\textbf{+3.5}}}$ & 30.0$_{\textcolor{darkgray}{\textbf{+5.0}}}$ & \underline{72.0}$_{\textcolor{darkgray}{\textbf{+5.7}}}$ & - \\
        \pretrain PPT Unsup.~\cite{wu2024towards} & CVPR 2024 & \scannet, \structure, \sdis & Contrast & \underline{75.8}$_{\textcolor{darkgray}{\textbf{+3.6}}}$ & 30.4$_{\textcolor{darkgray}{\textbf{+5.4}}}$ & 71.9$_{\textcolor{darkgray}{\textbf{+5.6}}}$ & - \\
        \rowcolor{ourscolor} \pretrain \textbf{\ours} & - & \scannet & Rendering & \textbf{76.0}$_{\textcolor{offsetcolor}{\textbf{+3.8}}}$ & \textbf{34.3}$_{\textcolor{offsetcolor}{\textbf{+9.3}}}$ & \textbf{72.1}$_{\textcolor{offsetcolor}{\textbf{+5.8}}}$ & \textbf{76.8}$_{\textcolor{offsetcolor}{\textbf{+4.4}}}$ \\
        \midrule
        \multicolumn{8}{c}{\textcolor{darkgray}{\textit{Supervised Pre-training}}} \\
        \midrule
        \textcolor{darkgray}{\suppretrain PPT Sup.~\cite{wu2024towards}} & \textcolor{darkgray}{CVPR 2024} & \textcolor{darkgray}{\scannet, \structure, \sdis} & \textcolor{darkgray}{3D Sup.} & \textcolor{darkgray}{76.4$_{\textcolor{darkgray}{\textbf{+4.2}}}$} & \textcolor{darkgray}{31.9$_{\textcolor{darkgray}{\textbf{+6.9}}}$} & \textcolor{darkgray}{72.7$_{\textcolor{darkgray}{\textbf{+6.4}}}$} & \textcolor{darkgray}{78.1$_{\textcolor{darkgray}{\textbf{+5.7}}}$} \\
        \textcolor{darkgray}{\suppretrain PonderV2~\cite{zhu2023ponderv2}} & \textcolor{darkgray}{arXiv 2024} & \textcolor{darkgray}{\scannet, \structure, \sdis} & \textcolor{darkgray}{2D Sup.} & \textcolor{darkgray}{77.0$_{\textcolor{darkgray}{\textbf{+4.8}}}$} & \textcolor{darkgray}{32.3$_{\textcolor{darkgray}{\textbf{+7.3}}}$} & \textcolor{darkgray}{73.2$_{\textcolor{darkgray}{\textbf{+6.9}}}$} & \textcolor{darkgray}{79.9$_{\textcolor{darkgray}{\textbf{+7.4}}}$} \\
        \textcolor{darkgray}{\suppretrain ARKit LM~\cite{ji2024arkit}} & \textcolor{darkgray}{CVPR 2025} & \textcolor{darkgray}{ALS200, \scannet/\scannetp} & \textcolor{darkgray}{3D Sup.} & \textcolor{darkgray}{77.0$_{\textcolor{darkgray}{\textbf{+4.8}}}$} & \textcolor{darkgray}{30.6$_{\textcolor{darkgray}{\textbf{+5.6}}}$} & \textcolor{darkgray}{-} & \textcolor{darkgray}{-} \\
        \bottomrule
    \end{tabular}
}
    \label{tab:sem_seg}
\end{table*}

\subsubsection{\textbf{3D Semantic Segmentation}}
In~\tabref{tab:sem_seg}, we present mIoU (\%) results for 3D semantic segmentation on \scannet~\cite{dai2017scannet}, \scannetp~\cite{rozenberszki2022lground}, and \sdis~\cite{armeni2016s3dis} benchmarks. Under the self-supervised pre-training setting (\cf \pretrain), \ours attains the best performance across all datasets, demonstrating a 76.0\% mIoU on \scannet validation set - a 2.5\% absolute improvement over prior neural rendering approaches such as GS$^3$~\cite{liu2024point} and Ponder~\cite{huang2023ponder}. Moreover, our method outperforms multi-datasets pre-training strategies MSC~\cite{wu2023masked} and PPT Unsup.~\cite{wu2024towards} by 4.8\% and 3.2\% on \scannetp, respectively. Although supervised pre-training baselines (\cf \suppretrain) maintain marginal advantages on \scannet ($\leq$1\%), our method establishes new state-of-the-art on \scannetp by enhanced semantic discriminability. This demonstrates the generalization of our method in learning transferable 3D representations and the potential of processing semantically complex scenarios. Consistent performance gains are observed on \sdis under both Area5 (72.1\%) and 6-fold cross-validation (76.8\%) settings, confirming its robustness.

\begin{table}[t]
    \centering
    \caption{3D instance segmentation performance on \scannet~\cite{dai2017scannet} and \scannetp~\cite{rozenberszki2022lground}. \textit{PG} indicates PointGroup~\cite{jiang2020pointgroup}.}
    \tablestyle{1.5pt}{1.0}
    
\resizebox{0.48\textwidth}{!}{
    \begin{tabular}{y{12mm}x{8mm}x{8mm}x{8mm}x{8mm}x{8mm}x{8mm}}
        \toprule
        \textbf{\textit{Ins. Seg.}} & \multicolumn{3}{c}{\scannet} & \multicolumn{3}{c}{\scannetp} \\
        \cmidrule(lr){1-1} \cmidrule(lr){2-4} \cmidrule(lr){5-7}
        Methods & AP$_{25}$ & AP$_{50}$ & mAP & AP$_{25}$ & AP$_{50}$ & mAP \\
        \midrule
        \rowcolor{gray} \scratch PG~\cite{jiang2020pointgroup} & {72.8} & {56.9} & {36.0} & {32.2} & {24.5} & {15.8} \\
        \midrule
        \pretrain PC~\cite{xie2020pointcontrast} & - & 58.0 & - & - & 24.9 & - \\
        \pretrain GS$^3$~\cite{liu2024point} & - & 59.2 & 37.0 & - & - & - \\
        \pretrain CSC~\cite{hou2021csc} & - & 59.4 & - & - & 25.2 & - \\
        \pretrain MSC~\cite{wu2023masked} & 74.7 & 59.6 & 39.3 & 34.3 & 26.8 & 17.3 \\
        \pretrain GC~\cite{wang2024groupcontrast} & - & {62.3} & - & - & {27.5} & - \\
        \midrule
        \rowcolor{ourscolor} \pretrain \textbf{Ours} & \textbf{77.0}$_{\textcolor{offsetcolor}{\textbf{+4.2}}}$ & \textbf{62.7}$_{\textcolor{offsetcolor}{\textbf{+6.2}}}$ & \textbf{40.8}$_{\textcolor{offsetcolor}{\textbf{+4.8}}}$ & \textbf{38.4}$_{\textcolor{offsetcolor}{\textbf{+5.8}}}$ & \textbf{30.6}$_{\textcolor{offsetcolor}{\textbf{+6.1}}}$ & \textbf{20.6}$_{\textcolor{offsetcolor}{\textbf{+4.8}}}$ \\
        \bottomrule
    \end{tabular}
}
    \label{tab:ins_seg}
\end{table}

\begin{figure}
    \centering
    \includegraphics[width=0.99\linewidth]{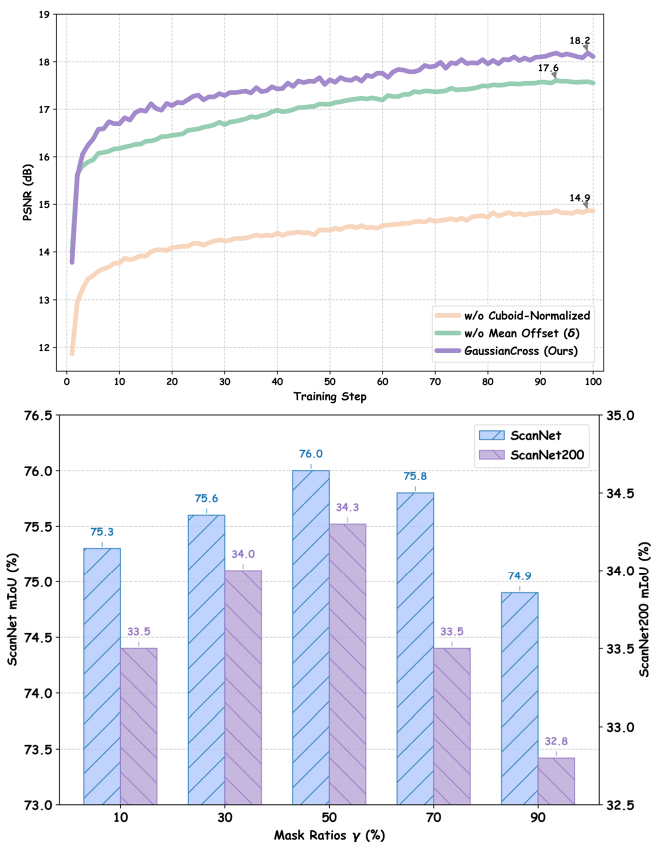}
    \caption{Ablation study of core designs and masking ratio $\gamma$.}
    \label{fig:ablation}
\end{figure}

\subsubsection{\textbf{3D Instance Segmentation}}
In~\tabref{tab:ins_seg}, we compare the results of instance segmentation on \scannet~\cite{dai2017scannet} and \scannetp~\cite{rozenberszki2022lground} validation splits with PointGroup~\cite{jiang2020pointgroup} as the baseline model. We report AP$_{25}$, AP$_{50}$, and mAP for comprehensive evaluation, following the common practice~\cite{jiang2020pointgroup,yao2024sgiformer}. On \scannet, the achieved 62.7\% AP$_{50}$ represents a 6.2\% improvement over the baseline without pre-training, significantly outperforming previous contrastive learning methods that typically struggle with instance boundary discrimination. The performance gap is more pronounced on \scannetp, where \ours attains 30.6\% mAP. The consistent superiority suggests that our method provides complementary benefits beyond pure color rendering (GS$^3$), underscoring the effectiveness of our designs in instance-level understanding.

\subsection{Ablation Studies and Analysis}
We perform systematic ablation studies to investigate the efficacy of our core designs and analyze the effect of different parameter choices. We utilize 3D semantic segmentation and assess the performance on both \scannet~\cite{dai2017scannet} and \scannetp~\cite{rozenberszki2022lground} validation splits for a comprehensive evaluation.

\begin{table}[t]
    \centering
    \caption{Ablation study of rendering targets. \textit{img., dep., sem.} denote RGB image, depth, and semantic feature maps.}
    \tablestyle{1.5pt}{1.0}
    \resizebox{0.45\textwidth}{!}{
    \begin{tabular}{x{8mm}x{8mm}x{8mm}x{7mm}x{7mm}x{7mm}x{7mm}}
        \toprule
        \multirow{2}{*}{\textit{w/ img.}} & \multirow{2}{*}{\textit{w/ dep.}} & \multirow{2}{*}{\textit{w/ sem.}} & \multicolumn{2}{c}{\scannet} & \multicolumn{2}{c}{\scannetp} \\
        \cmidrule(lr){4-5} \cmidrule(lr){6-7}
        {} & {} & {} & mIoU & mAcc & mIoU & mAcc \\
        \midrule
        \YesV & \NoX & \NoX & 75.0 & 82.9 & 32.8 & 42.1 \\
        \YesV & \YesV & \NoX & 75.3 & 83.0 & 33.0 & 42.4 \\
        \YesV & \NoX & \YesV & 75.5 & 83.0 & 33.7 & 42.5 \\
        \rowcolor{ourscolor} \YesV & \YesV & \YesV & \textbf{76.0} & \textbf{83.5} & \textbf{34.3} & \textbf{43.1} \\
        \bottomrule
    \end{tabular}
}
    \label{tab:ablation_target}
\end{table}

\mypara{Core Designs.} 
In~\figref{fig:ablation} top, we analyze the impact of our core designs by recording the PSNR of rendered images during pre-training. We observe that using traditional Gaussian mean initialization leads to a significant drop (14.9 \textit{v.s.} 18.2), indicating that the model struggles to learn meaningful representations. The variant without Gaussian mean refinement achieves a PSNR of 17.6, suggesting that the learned offset can help with accurate scene representation. Different rendering targets specialize in distinct attributes of 3D scenes, thus impacting the representations. Therefore, we explore the synergistic effects of multi-target rendering in~\tabref{tab:ablation_target}. The baseline using only photometric reconstruction achieves 75.0\% mIoU on \scannet and 32.8\% on \scannetp, establishing a performance floor that highlights the limitation of pure appearance modeling. Incorporating geometric consistency by depth supervision yields a slight improvement, revealing that explicit spatial cues enhance 3D structure understanding. The performance is elevated to 75.5\% and 33.7\% when bridging semantic alignment via knowledge distillation. The optimal configuration combining photometric, geometric, and semantic targets achieves 76.0\% and 34.1\% mIoU, respectively, proving the complementary nature of tripartite rendering.

\mypara{Masking Ratio $\gamma$.} 
We adopt a stochastic masking strategy governed by parameter $\gamma$ to occlude a portion of input regions during pre-training. To test its impact, we vary $\gamma$ from 10\% to 90\% in 20\% increments. As evidenced in~\figref{fig:ablation}, the results show that better performance can be achieved when $\gamma$ equals 50\%, with perturbations within ±20\% causing statistically insignificant performance deviations. However, extreme values of 10\% or 90\% induce significant performance degradation, revealing the model's sensitivity to excessive occlusion or exposure. This suggests the importance of balanced masking in self-supervised learning.

\begin{table}[t]
    \centering
    \caption{Impact of opacity threshold $\tau$.}
    \tablestyle{1.5pt}{1.0}
    
\resizebox{0.48\textwidth}{!}{
    \begin{tabular}{x{6mm}|x{7mm}x{8mm}x{8mm}|x{8mm}x{9mm}x{8mm}}
        \toprule
        \multirow{2}{*}{$\tau$} & Sc. & Sc.200 & Average & PSNR$\uparrow$ & Memory$\downarrow$ & Time$\downarrow$ \\
        {} & mIoU & mIoU & mIoU & (dB) & (MB) & (s/scene) \\
        \midrule
        0.1 & 75.4 & 33.3 & 54.3 & 18.16 & 5614 & 0.265 \\
        \rowcolor{ourscolor} 0.3 & \textbf{76.0} & \textbf{34.3} & \textbf{55.1} & \textbf{18.18} & 5296 & 0.255 \\
        0.5 & 75.6 & 33.6 & 54.6 & 17.94 & 5251 & 0.251 \\
        0.7 & 74.8 & 33.0 & 53.9 & 17.62 & \textbf{5204} & \textbf{0.241} \\
        \bottomrule
    \end{tabular}
}
    \label{tab:ablation_opacity}
\end{table}

\mypara{Opacity Threshold $\tau$.} 
We introduce an opacity-driven pruning strategy to determine the visibility of each anchor Gaussian and optimize the rendering quality. In~\tabref{tab:ablation_opacity}, we examine $\tau$ from 0.1 to 0.7. We also report memory consumption and training time for each scene. When increasing the threshold from 0.1 to 0.3, the performance is also improved, while further raising the value to 0.5 or 0.7 will lead to a drop. This is because a higher threshold will filter out more anchors. Therefore, we set $\tau$ to 0.3 in our experiments to balance rendering quality and amount of information.

\begin{table}[t]
    \centering
    \caption{Effectiveness of $M$ and $\gX_{f}$.}
    \tablestyle{1.5pt}{1.0}
    \resizebox{0.48\textwidth}{!}{
    \begin{tabular}{y{12mm}x{8mm}|x{6mm}x{6mm}x{6mm}|x{6mm}x{8mm}x{8mm}}
        \toprule
        \multicolumn{2}{c}{\multirow{2}{*}{\textit{Datasets \& Metrics}}} & \multicolumn{3}{c}{Rendering Views $M$} & \multicolumn{3}{c}{VFMs $\gX_{f}$} \\
        \cmidrule(lr){3-5} \cmidrule(lr){6-8}
        \multicolumn{2}{c}{} & 2 & \cellcolor{ourscolor} 5 & 8 & CLIP & DINOv2 & \cellcolor{ourscolor} RADIO \\
        \midrule
        \multirow{2}{*}{\scannet} & mIoU & 75.8 & \cellcolor{ourscolor} \textbf{76.0} & 75.6 & 75.4 & 74.9 & \cellcolor{ourscolor} \textbf{76.0} \\
        {} & mAcc & \textbf{83.7} & \cellcolor{ourscolor} {83.5} & 83.5 & 83.5 & 83.1 & \cellcolor{ourscolor} 83.5 \\
        \midrule
        \multirow{2}{*}{\scannetp} & mIoU & {33.9} & \cellcolor{ourscolor} \textbf{34.3} & 34.0 & 33.9 & 33.6 & \cellcolor{ourscolor} \textbf{34.3} \\
        {} & mAcc & 42.8 & \cellcolor{ourscolor} \textbf{43.1} & {42.9} & 42.9 & 43.0 & \cellcolor{ourscolor} \textbf{43.1} \\
        \midrule
        \multirow{2}{*}{Average} & mIoU & 54.3 & \cellcolor{ourscolor} \textbf{55.1} & 54.6 & 54.6 & 54.2 & \cellcolor{ourscolor} \textbf{55.1} \\
        {} & mAcc & 62.4 & \cellcolor{ourscolor} \textbf{63.3} & 62.1 & 63.2 & 63.0 & \cellcolor{ourscolor} \textbf{63.3} \\
        \bottomrule
    \end{tabular}
}
    \label{tab:ablation_view}
\end{table}

\mypara{Visual Foundation Models $\gX_{f}$.} 
\ours's architectural flexibility allows for seamless integration with diverse visual foundation models. However, different models excel at distinctive properties that affect scene understanding. Results in~\tabref{tab:ablation_opacity} indicate notable performance variance across foundation models, with CLIP~\cite{radford2021learning} and DINOv2~\cite{oquab2023dinov2} yielding suboptimal results. Because of the agglomerative multi-domain training strategy, RADIO~\cite{heinrich2025radiov25} achieves optimal 76.0\% mIoU on \scannet and 34.1\% on \scannetp.

\mypara{Number of Rendering Views $M$.} 
Theoretically, more views could offer broader supervision for pre-training, but it also introduce extra computational costs and increase training time. Thus, we investigate the impact of $M$ in~\tabref{tab:ablation_view}. We set $M$ to 5 in our experiments to balance the performance and efficiency.

\section{Conclusion}
\label{sec:conclusion}
In this paper, we present \ours, an innovative framework leveraging 3DGS for cross-modal self-supervised point cloud representation learning. Our cuboid-normalized Gaussian initialization establishes scale-consistent scene representations by transforming raw point clouds into a structured collection of Gaussian primitives within a canonical space. The proposed tri-attribute adaptive distillation splatting jointly optimizes photometric appearance, geometric structure, and semantic consistency by differentiable rendering with a feature field while effectively distilling the 2D visual foundation model for enhanced semantic awareness. Extensive experiments demonstrate state-of-the-art performance across multiple benchmarks, including linear probing and transfer learning. Comprehensive ablation studies further validate the effectiveness by systematically analyzing core design components. For future work, we will explore scalable backbone architectures to enhance representation capability and investigate the potential of scaling up \ours to large-scale multi-source datasets, aiming to advance the development of 3D foundation models.

\begin{acks}
The research work was conducted in the JC STEM Lab of Machine Learning and Computer Vision funded by The Hong Kong Jockey Club Charities Trust. 
\end{acks}

\bibliographystyle{ACM-Reference-Format}
\balance
\bibliography{references}


\begin{thebibliography}{52}


\ifx \showCODEN    \undefined \def \showCODEN     #1{\unskip}     \fi
\ifx \showDOI      \undefined \def \showDOI       #1{#1}\fi
\ifx \showISBNx    \undefined \def \showISBNx     #1{\unskip}     \fi
\ifx \showISBNxiii \undefined \def \showISBNxiii  #1{\unskip}     \fi
\ifx \showISSN     \undefined \def \showISSN      #1{\unskip}     \fi
\ifx \showLCCN     \undefined \def \showLCCN      #1{\unskip}     \fi
\ifx \shownote     \undefined \def \shownote      #1{#1}          \fi
\ifx \showarticletitle \undefined \def \showarticletitle #1{#1}   \fi
\ifx \showURL      \undefined \def \showURL       {\relax}        \fi
\providecommand\bibfield[2]{#2}
\providecommand\bibinfo[2]{#2}
\providecommand\natexlab[1]{#1}
\providecommand\showeprint[2][]{arXiv:#2}

\bibitem[Armeni et~al\mbox{.}(2016)]%
        {armeni2016s3dis}
\bibfield{author}{\bibinfo{person}{Iro Armeni}, \bibinfo{person}{Ozan Sener}, \bibinfo{person}{Amir~R Zamir}, \bibinfo{person}{Helen Jiang}, \bibinfo{person}{Ioannis Brilakis}, \bibinfo{person}{Martin Fischer}, {and} \bibinfo{person}{Silvio Savarese}.} \bibinfo{year}{2016}\natexlab{}.
\newblock \showarticletitle{3d semantic parsing of large-scale indoor spaces}. In \bibinfo{booktitle}{\emph{Proceedings of the IEEE/CVF Conference on Computer Vision and Pattern Recognition (CVPR)}}. \bibinfo{publisher}{IEEE}.
\newblock


\bibitem[Charatan et~al\mbox{.}(2024)]%
        {charatan2024pixelsplat}
\bibfield{author}{\bibinfo{person}{David Charatan}, \bibinfo{person}{Sizhe~Lester Li}, \bibinfo{person}{Andrea Tagliasacchi}, {and} \bibinfo{person}{Vincent Sitzmann}.} \bibinfo{year}{2024}\natexlab{}.
\newblock \showarticletitle{pixelsplat: 3d gaussian splats from image pairs for scalable generalizable 3d reconstruction}. In \bibinfo{booktitle}{\emph{Proceedings of the IEEE/CVF Conference on Computer Vision and Pattern Recognition (CVPR)}}. \bibinfo{publisher}{IEEE}, \bibinfo{pages}{19457--19467}.
\newblock


\bibitem[Chen et~al\mbox{.}(2024)]%
        {chen2024mvsplat}
\bibfield{author}{\bibinfo{person}{Yuedong Chen}, \bibinfo{person}{Haofei Xu}, \bibinfo{person}{Chuanxia Zheng}, \bibinfo{person}{Bohan Zhuang}, \bibinfo{person}{Marc Pollefeys}, \bibinfo{person}{Andreas Geiger}, \bibinfo{person}{Tat-Jen Cham}, {and} \bibinfo{person}{Jianfei Cai}.} \bibinfo{year}{2024}\natexlab{}.
\newblock \showarticletitle{Mvsplat: Efficient 3d gaussian splatting from sparse multi-view images}. In \bibinfo{booktitle}{\emph{European Conference on Computer Vision (ECCV)}}. \bibinfo{publisher}{Springer}, \bibinfo{pages}{370--386}.
\newblock


\bibitem[Choy et~al\mbox{.}(2019)]%
        {choy2019minkowski}
\bibfield{author}{\bibinfo{person}{Christopher Choy}, \bibinfo{person}{JunYoung Gwak}, {and} \bibinfo{person}{Silvio Savarese}.} \bibinfo{year}{2019}\natexlab{}.
\newblock \showarticletitle{4d spatio-temporal convnets: Minkowski convolutional neural networks}. In \bibinfo{booktitle}{\emph{Proceedings of the IEEE/CVF Conference on Computer Vision and Pattern Recognition (CVPR)}}. \bibinfo{publisher}{IEEE}.
\newblock


\bibitem[Contributors(2023)]%
        {pointcept2023}
\bibfield{author}{\bibinfo{person}{Pointcept Contributors}.} \bibinfo{year}{2023}\natexlab{}.
\newblock \bibinfo{title}{Pointcept: A Codebase for Point Cloud Perception Research}.
\newblock \bibinfo{howpublished}{\url{https://github.com/Pointcept/Pointcept}}.
\newblock


\bibitem[Dai et~al\mbox{.}(2017)]%
        {dai2017scannet}
\bibfield{author}{\bibinfo{person}{Angela Dai}, \bibinfo{person}{Angel~X Chang}, \bibinfo{person}{Manolis Savva}, \bibinfo{person}{Maciej Halber}, \bibinfo{person}{Thomas Funkhouser}, {and} \bibinfo{person}{Matthias Nie{\ss}ner}.} \bibinfo{year}{2017}\natexlab{}.
\newblock \showarticletitle{Scannet: Richly-annotated 3d reconstructions of indoor scenes}. In \bibinfo{booktitle}{\emph{Proceedings of the IEEE/CVF Conference on Computer Vision and Pattern Recognition (CVPR)}}. \bibinfo{publisher}{IEEE}.
\newblock


\bibitem[Devlin et~al\mbox{.}(2019)]%
        {devlin2019bert}
\bibfield{author}{\bibinfo{person}{Jacob Devlin}, \bibinfo{person}{Ming-Wei Chang}, \bibinfo{person}{Kenton Lee}, {and} \bibinfo{person}{Kristina Toutanova}.} \bibinfo{year}{2019}\natexlab{}.
\newblock \showarticletitle{Bert: Pre-training of deep bidirectional transformers for language understanding}. In \bibinfo{booktitle}{\emph{Proceedings of the 2019 conference of the North American chapter of the association for computational linguistics: human language technologies, volume 1 (long and short papers)}}. \bibinfo{publisher}{Association for Computational Linguistics}, \bibinfo{pages}{4171--4186}.
\newblock


\bibitem[Fan et~al\mbox{.}(2024)]%
        {fan2024point}
\bibfield{author}{\bibinfo{person}{Guofan Fan}, \bibinfo{person}{Zekun Qi}, \bibinfo{person}{Wenkai Shi}, {and} \bibinfo{person}{Kaisheng Ma}.} \bibinfo{year}{2024}\natexlab{}.
\newblock \showarticletitle{Point-gcc: Universal self-supervised 3d scene pre-training via geometry-color contrast}. In \bibinfo{booktitle}{\emph{Proceedings of the 32nd ACM International Conference on Multimedia}}. \bibinfo{pages}{4709--4718}.
\newblock


\bibitem[Felzenszwalb and Huttenlocher(2004)]%
        {felzenszwalb2004efficient}
\bibfield{author}{\bibinfo{person}{Pedro~F Felzenszwalb} {and} \bibinfo{person}{Daniel~P Huttenlocher}.} \bibinfo{year}{2004}\natexlab{}.
\newblock \showarticletitle{Efficient graph-based image segmentation}.
\newblock \bibinfo{journal}{\emph{International Journal of Computer Vision}}  \bibinfo{volume}{59} (\bibinfo{year}{2004}), \bibinfo{pages}{167--181}.
\newblock


\bibitem[Graham and Van~der Maaten(2017)]%
        {graham2017submanifold}
\bibfield{author}{\bibinfo{person}{Benjamin Graham} {and} \bibinfo{person}{Laurens Van~der Maaten}.} \bibinfo{year}{2017}\natexlab{}.
\newblock \showarticletitle{Submanifold sparse convolutional networks}.
\newblock \bibinfo{journal}{\emph{arXiv preprint arXiv:1706.01307}} (\bibinfo{year}{2017}).
\newblock


\bibitem[He et~al\mbox{.}(2022)]%
        {he2022masked}
\bibfield{author}{\bibinfo{person}{Kaiming He}, \bibinfo{person}{Xinlei Chen}, \bibinfo{person}{Saining Xie}, \bibinfo{person}{Yanghao Li}, \bibinfo{person}{Piotr Doll{\'a}r}, {and} \bibinfo{person}{Ross Girshick}.} \bibinfo{year}{2022}\natexlab{}.
\newblock \showarticletitle{Masked autoencoders are scalable vision learners}. In \bibinfo{booktitle}{\emph{Proceedings of the IEEE/CVF International Conference on Computer Vision (ICCV)}}. \bibinfo{publisher}{IEEE}, \bibinfo{pages}{16000--16009}.
\newblock


\bibitem[He et~al\mbox{.}(2020)]%
        {he2020momentum}
\bibfield{author}{\bibinfo{person}{Kaiming He}, \bibinfo{person}{Haoqi Fan}, \bibinfo{person}{Yuxin Wu}, \bibinfo{person}{Saining Xie}, {and} \bibinfo{person}{Ross Girshick}.} \bibinfo{year}{2020}\natexlab{}.
\newblock \showarticletitle{Momentum contrast for unsupervised visual representation learning}. In \bibinfo{booktitle}{\emph{Proceedings of the IEEE/CVF conference on computer vision and pattern recognition}}. \bibinfo{pages}{9729--9738}.
\newblock


\bibitem[Heinrich et~al\mbox{.}(2025)]%
        {heinrich2025radiov25}
\bibfield{author}{\bibinfo{person}{Greg Heinrich}, \bibinfo{person}{Mike Ranzinger}, \bibinfo{person}{Hongxu}, \bibinfo{person}{Yin}, \bibinfo{person}{Yao Lu}, \bibinfo{person}{Jan Kautz}, \bibinfo{person}{Andrew Tao}, \bibinfo{person}{Bryan Catanzaro}, {and} \bibinfo{person}{Pavlo Molchanov}.} \bibinfo{year}{2025}\natexlab{}.
\newblock \showarticletitle{RADIOv2.5: Improved Baselines for Agglomerative Vision Foundation Models}. In \bibinfo{booktitle}{\emph{Proceedings of the IEEE/CVF Conference on Computer Vision and Pattern Recognition}}.
\newblock


\bibitem[Hou et~al\mbox{.}(2021)]%
        {hou2021csc}
\bibfield{author}{\bibinfo{person}{Ji Hou}, \bibinfo{person}{Benjamin Graham}, \bibinfo{person}{Matthias Nie{\ss}ner}, {and} \bibinfo{person}{Saining Xie}.} \bibinfo{year}{2021}\natexlab{}.
\newblock \showarticletitle{Exploring data-efficient 3d scene understanding with contrastive scene contexts}. In \bibinfo{booktitle}{\emph{Proceedings of the IEEE/CVF Conference on Computer Vision and Pattern Recognition (CVPR)}}. \bibinfo{publisher}{IEEE}, \bibinfo{pages}{15587--15597}.
\newblock


\bibitem[Huang et~al\mbox{.}(2023)]%
        {huang2023ponder}
\bibfield{author}{\bibinfo{person}{Di Huang}, \bibinfo{person}{Sida Peng}, \bibinfo{person}{Tong He}, \bibinfo{person}{Honghui Yang}, \bibinfo{person}{Xiaowei Zhou}, {and} \bibinfo{person}{Wanli Ouyang}.} \bibinfo{year}{2023}\natexlab{}.
\newblock \showarticletitle{Ponder: Point cloud pre-training via neural rendering}. In \bibinfo{booktitle}{\emph{Proceedings of the IEEE/CVF International Conference on Computer Vision (ICCV)}}. \bibinfo{publisher}{IEEE}, \bibinfo{pages}{16089--16098}.
\newblock


\bibitem[Ji et~al\mbox{.}(2024)]%
        {ji2024arkit}
\bibfield{author}{\bibinfo{person}{Guangda Ji}, \bibinfo{person}{Silvan Weder}, \bibinfo{person}{Francis Engelmann}, \bibinfo{person}{Marc Pollefeys}, {and} \bibinfo{person}{Hermann Blum}.} \bibinfo{year}{2024}\natexlab{}.
\newblock \showarticletitle{ARKit LabelMaker: A New Scale for Indoor 3D Scene Understanding}.
\newblock \bibinfo{journal}{\emph{arXiv preprint arXiv:2410.13924}} (\bibinfo{year}{2024}).
\newblock


\bibitem[Jiang et~al\mbox{.}(2020)]%
        {jiang2020pointgroup}
\bibfield{author}{\bibinfo{person}{Li Jiang}, \bibinfo{person}{Hengshuang Zhao}, \bibinfo{person}{Shaoshuai Shi}, \bibinfo{person}{Shu Liu}, \bibinfo{person}{Chi-Wing Fu}, {and} \bibinfo{person}{Jiaya Jia}.} \bibinfo{year}{2020}\natexlab{}.
\newblock \showarticletitle{Pointgroup: Dual-set point grouping for 3d instance segmentation}. In \bibinfo{booktitle}{\emph{Proceedings of the IEEE/CVF Conference on Computer Vision and Pattern Recognition (CVPR)}}. \bibinfo{publisher}{IEEE}, \bibinfo{pages}{4867--4876}.
\newblock


\bibitem[Kerbl et~al\mbox{.}(2023)]%
        {kerbl20233d}
\bibfield{author}{\bibinfo{person}{Bernhard Kerbl}, \bibinfo{person}{Georgios Kopanas}, \bibinfo{person}{Thomas Leimk{\"u}hler}, {and} \bibinfo{person}{George Drettakis}.} \bibinfo{year}{2023}\natexlab{}.
\newblock \showarticletitle{3d gaussian splatting for real-time radiance field rendering.}
\newblock \bibinfo{journal}{\emph{ACM TOG}} \bibinfo{volume}{42}, \bibinfo{number}{4} (\bibinfo{year}{2023}), \bibinfo{pages}{139--1}.
\newblock


\bibitem[Lai et~al\mbox{.}(2022)]%
        {lai2022stratified}
\bibfield{author}{\bibinfo{person}{Xin Lai}, \bibinfo{person}{Jianhui Liu}, \bibinfo{person}{Li Jiang}, \bibinfo{person}{Liwei Wang}, \bibinfo{person}{Hengshuang Zhao}, \bibinfo{person}{Shu Liu}, \bibinfo{person}{Xiaojuan Qi}, {and} \bibinfo{person}{Jiaya Jia}.} \bibinfo{year}{2022}\natexlab{}.
\newblock \showarticletitle{Stratified transformer for 3d point cloud segmentation}. In \bibinfo{booktitle}{\emph{Proceedings of the IEEE/CVF Conference on Computer Vision and Pattern Recognition (CVPR)}}. \bibinfo{publisher}{IEEE}.
\newblock


\bibitem[Lemke et~al\mbox{.}(2024)]%
        {zurbrugg2024icgnet}
\bibfield{author}{\bibinfo{person}{Oliver Lemke}, \bibinfo{person}{Zuria Bauer}, \bibinfo{person}{Ren{\'e} Zurbr{\"u}gg}, \bibinfo{person}{Marc Pollefeys}, \bibinfo{person}{Francis Engelmann}, {and} \bibinfo{person}{Hermann Blum}.} \bibinfo{year}{2024}\natexlab{}.
\newblock \showarticletitle{{Spot-Compose: A Framework for Open-Vocabulary Object Retrieval and Drawer Manipulation in Point Clouds}}.
\newblock \bibinfo{journal}{\emph{Internationl Conference on Robotics and Automation Workshops (ICRAW)}} (\bibinfo{year}{2024}).
\newblock


\bibitem[Liu et~al\mbox{.}(2024a)]%
        {liu2024point}
\bibfield{author}{\bibinfo{person}{Hao Liu}, \bibinfo{person}{Minglin Chen}, \bibinfo{person}{Yanni Ma}, \bibinfo{person}{Haihong Xiao}, {and} \bibinfo{person}{Ying He}.} \bibinfo{year}{2024}\natexlab{a}.
\newblock \showarticletitle{Point Cloud Unsupervised Pre-training via 3D Gaussian Splatting}.
\newblock \bibinfo{journal}{\emph{arXiv preprint arXiv:2411.18667}} (\bibinfo{year}{2024}).
\newblock


\bibitem[Liu et~al\mbox{.}(2024b)]%
        {liu2024menet}
\bibfield{author}{\bibinfo{person}{Moyun Liu}, \bibinfo{person}{Youping Chen}, \bibinfo{person}{Jingming Xie}, \bibinfo{person}{Yijie Zhu}, \bibinfo{person}{Yang Zhang}, \bibinfo{person}{Lei Yao}, \bibinfo{person}{Zhenshan Bing}, \bibinfo{person}{Genghang Zhuang}, \bibinfo{person}{Kai Huang}, {and} \bibinfo{person}{Joey~Tianyi Zhou}.} \bibinfo{year}{2024}\natexlab{b}.
\newblock \showarticletitle{MENet: Multi-modal mapping enhancement network for 3D object detection in autonomous driving}.
\newblock \bibinfo{journal}{\emph{IEEE Transactions on Intelligent Transportation Systems}} \bibinfo{volume}{25}, \bibinfo{number}{8} (\bibinfo{year}{2024}), \bibinfo{pages}{9397--9410}.
\newblock


\bibitem[McInnes et~al\mbox{.}(2018)]%
        {mcinnes2018umap}
\bibfield{author}{\bibinfo{person}{Leland McInnes}, \bibinfo{person}{John Healy}, {and} \bibinfo{person}{James Melville}.} \bibinfo{year}{2018}\natexlab{}.
\newblock \showarticletitle{Umap: Uniform manifold approximation and projection for dimension reduction}.
\newblock \bibinfo{journal}{\emph{arXiv preprint arXiv:1802.03426}} (\bibinfo{year}{2018}).
\newblock


\bibitem[Mildenhall et~al\mbox{.}(2021)]%
        {mildenhall2021nerf}
\bibfield{author}{\bibinfo{person}{Ben Mildenhall}, \bibinfo{person}{Pratul~P Srinivasan}, \bibinfo{person}{Matthew Tancik}, \bibinfo{person}{Jonathan~T Barron}, \bibinfo{person}{Ravi Ramamoorthi}, {and} \bibinfo{person}{Ren Ng}.} \bibinfo{year}{2021}\natexlab{}.
\newblock \showarticletitle{Nerf: Representing scenes as neural radiance fields for view synthesis}.
\newblock \bibinfo{journal}{\emph{Commun. ACM}} \bibinfo{volume}{65}, \bibinfo{number}{1} (\bibinfo{year}{2021}), \bibinfo{pages}{99--106}.
\newblock


\bibitem[Oquab et~al\mbox{.}(2023)]%
        {oquab2023dinov2}
\bibfield{author}{\bibinfo{person}{Maxime Oquab}, \bibinfo{person}{Timoth{\'e}e Darcet}, \bibinfo{person}{Th{\'e}o Moutakanni}, \bibinfo{person}{Huy Vo}, \bibinfo{person}{Marc Szafraniec}, \bibinfo{person}{Vasil Khalidov}, \bibinfo{person}{Pierre Fernandez}, \bibinfo{person}{Daniel Haziza}, \bibinfo{person}{Francisco Massa}, \bibinfo{person}{Alaaeldin El-Nouby}, {et~al\mbox{.}}} \bibinfo{year}{2023}\natexlab{}.
\newblock \showarticletitle{Dinov2: Learning robust visual features without supervision}.
\newblock \bibinfo{journal}{\emph{arXiv preprint arXiv:2304.07193}} (\bibinfo{year}{2023}).
\newblock


\bibitem[Pang et~al\mbox{.}(2022)]%
        {pang2022masked}
\bibfield{author}{\bibinfo{person}{Yatian Pang}, \bibinfo{person}{Wenxiao Wang}, \bibinfo{person}{Francis~EH Tay}, \bibinfo{person}{Wei Liu}, \bibinfo{person}{Yonghong Tian}, {and} \bibinfo{person}{Li Yuan}.} \bibinfo{year}{2022}\natexlab{}.
\newblock \showarticletitle{Masked autoencoders for point cloud self-supervised learning}. In \bibinfo{booktitle}{\emph{European Conference on Computer Vision (ECCV)}}, Vol.~\bibinfo{volume}{13662}. \bibinfo{publisher}{Springer}, \bibinfo{pages}{604--621}.
\newblock


\bibitem[Qi et~al\mbox{.}(2023)]%
        {qi2023contrast}
\bibfield{author}{\bibinfo{person}{Zekun Qi}, \bibinfo{person}{Runpei Dong}, \bibinfo{person}{Guofan Fan}, \bibinfo{person}{Zheng Ge}, \bibinfo{person}{Xiangyu Zhang}, \bibinfo{person}{Kaisheng Ma}, {and} \bibinfo{person}{Li Yi}.} \bibinfo{year}{2023}\natexlab{}.
\newblock \showarticletitle{Contrast with reconstruct: Contrastive 3d representation learning guided by generative pretraining}. In \bibinfo{booktitle}{\emph{International Conference on Machine Learning}}. PMLR, \bibinfo{pages}{28223--28243}.
\newblock


\bibitem[Qian et~al\mbox{.}(2022)]%
        {qian2022pointnext}
\bibfield{author}{\bibinfo{person}{Guocheng Qian}, \bibinfo{person}{Yuchen Li}, \bibinfo{person}{Houwen Peng}, \bibinfo{person}{Jinjie Mai}, \bibinfo{person}{Hasan Hammoud}, \bibinfo{person}{Mohamed Elhoseiny}, {and} \bibinfo{person}{Bernard Ghanem}.} \bibinfo{year}{2022}\natexlab{}.
\newblock \showarticletitle{Pointnext: Revisiting pointnet++ with improved training and scaling strategies}.
\newblock \bibinfo{journal}{\emph{Advances in Neural Information Processing Systems (NeurIPS)}} (\bibinfo{year}{2022}).
\newblock


\bibitem[Radford et~al\mbox{.}(2021)]%
        {radford2021learning}
\bibfield{author}{\bibinfo{person}{Alec Radford}, \bibinfo{person}{Jong~Wook Kim}, \bibinfo{person}{Chris Hallacy}, \bibinfo{person}{Aditya Ramesh}, \bibinfo{person}{Gabriel Goh}, \bibinfo{person}{Sandhini Agarwal}, \bibinfo{person}{Girish Sastry}, \bibinfo{person}{Amanda Askell}, \bibinfo{person}{Pamela Mishkin}, \bibinfo{person}{Jack Clark}, {et~al\mbox{.}}} \bibinfo{year}{2021}\natexlab{}.
\newblock \showarticletitle{Learning transferable visual models from natural language supervision}. In \bibinfo{booktitle}{\emph{International conference on machine learning}}. \bibinfo{pages}{8748--8763}.
\newblock


\bibitem[Rozenberszki et~al\mbox{.}(2022)]%
        {rozenberszki2022lground}
\bibfield{author}{\bibinfo{person}{David Rozenberszki}, \bibinfo{person}{Or Litany}, {and} \bibinfo{person}{Angela Dai}.} \bibinfo{year}{2022}\natexlab{}.
\newblock \showarticletitle{Language-grounded indoor 3d semantic segmentation in the wild}. In \bibinfo{booktitle}{\emph{European Conference on Computer Vision (ECCV)}}. Springer, \bibinfo{pages}{125--141}.
\newblock


\bibitem[Snavely et~al\mbox{.}(2006)]%
        {snavely2006photo}
\bibfield{author}{\bibinfo{person}{Noah Snavely}, \bibinfo{person}{Steven~M Seitz}, {and} \bibinfo{person}{Richard Szeliski}.} \bibinfo{year}{2006}\natexlab{}.
\newblock \showarticletitle{Photo tourism: exploring photo collections in 3D}.
\newblock In \bibinfo{booktitle}{\emph{ACM siggraph 2006 papers}}. Vol.~\bibinfo{volume}{25}. \bibinfo{pages}{835--846}.
\newblock


\bibitem[Vaswani et~al\mbox{.}(2017)]%
        {vaswani2017attention}
\bibfield{author}{\bibinfo{person}{Ashish Vaswani}, \bibinfo{person}{Noam Shazeer}, \bibinfo{person}{Niki Parmar}, \bibinfo{person}{Jakob Uszkoreit}, \bibinfo{person}{Llion Jones}, \bibinfo{person}{Aidan~N Gomez}, \bibinfo{person}{{\L}ukasz Kaiser}, {and} \bibinfo{person}{Illia Polosukhin}.} \bibinfo{year}{2017}\natexlab{}.
\newblock \showarticletitle{Attention is all you need}. In \bibinfo{booktitle}{\emph{Advances in Neural Information Processing Systems (NeurIPS)}}. \bibinfo{pages}{5998--6008}.
\newblock


\bibitem[Wang et~al\mbox{.}(2024b)]%
        {wang2024groupcontrast}
\bibfield{author}{\bibinfo{person}{Chengyao Wang}, \bibinfo{person}{Li Jiang}, \bibinfo{person}{Xiaoyang Wu}, \bibinfo{person}{Zhuotao Tian}, \bibinfo{person}{Bohao Peng}, \bibinfo{person}{Hengshuang Zhao}, {and} \bibinfo{person}{Jiaya Jia}.} \bibinfo{year}{2024}\natexlab{b}.
\newblock \showarticletitle{Groupcontrast: Semantic-aware self-supervised representation learning for 3d understanding}. In \bibinfo{booktitle}{\emph{Proceedings of the IEEE/CVF Conference on Computer Vision and Pattern Recognition (CVPR)}}. \bibinfo{publisher}{IEEE}, \bibinfo{pages}{4917--4928}.
\newblock


\bibitem[Wang et~al\mbox{.}(2024c)]%
        {wang2024pfgs}
\bibfield{author}{\bibinfo{person}{Jiaxu Wang}, \bibinfo{person}{Ziyi Zhang}, \bibinfo{person}{Junhao He}, {and} \bibinfo{person}{Renjing Xu}.} \bibinfo{year}{2024}\natexlab{c}.
\newblock \showarticletitle{PFGS: High Fidelity Point Cloud Rendering via Feature Splatting}. In \bibinfo{booktitle}{\emph{European Conference on Computer Vision}}. Springer, \bibinfo{pages}{193--209}.
\newblock


\bibitem[Wang et~al\mbox{.}(2021)]%
        {wang2021neus}
\bibfield{author}{\bibinfo{person}{Peng Wang}, \bibinfo{person}{Lingjie Liu}, \bibinfo{person}{Yuan Liu}, \bibinfo{person}{Christian Theobalt}, \bibinfo{person}{Taku Komura}, {and} \bibinfo{person}{Wenping Wang}.} \bibinfo{year}{2021}\natexlab{}.
\newblock \showarticletitle{Neus: Learning neural implicit surfaces by volume rendering for multi-view reconstruction}.
\newblock \bibinfo{journal}{\emph{arXiv preprint arXiv:2106.10689}} (\bibinfo{year}{2021}).
\newblock


\bibitem[Wang et~al\mbox{.}(2024a)]%
        {wang2024freesplat}
\bibfield{author}{\bibinfo{person}{Yunsong Wang}, \bibinfo{person}{Tianxin Huang}, \bibinfo{person}{Hanlin Chen}, {and} \bibinfo{person}{Gim~Hee Lee}.} \bibinfo{year}{2024}\natexlab{a}.
\newblock \showarticletitle{FreeSplat: Generalizable 3D Gaussian Splatting Towards Free-View Synthesis of Indoor Scenes}.
\newblock \bibinfo{journal}{\emph{Advances in Neural Information Processing Systems (NeurIPS)}} (\bibinfo{year}{2024}).
\newblock


\bibitem[Wu et~al\mbox{.}(2025)]%
        {wu2025sonata}
\bibfield{author}{\bibinfo{person}{Xiaoyang Wu}, \bibinfo{person}{Daniel DeTone}, \bibinfo{person}{Duncan Frost}, \bibinfo{person}{Tianwei Shen}, \bibinfo{person}{Chris Xie}, \bibinfo{person}{Nan Yang}, \bibinfo{person}{Jakob Engel}, \bibinfo{person}{Richard Newcombe}, \bibinfo{person}{Hengshuang Zhao}, {and} \bibinfo{person}{Julian Straub}.} \bibinfo{year}{2025}\natexlab{}.
\newblock \showarticletitle{Sonata: Self-Supervised Learning of Reliable Point Representations}. In \bibinfo{booktitle}{\emph{Proceedings of the IEEE/CVF Conference on Computer Vision and Pattern Recognition}}.
\newblock


\bibitem[Wu et~al\mbox{.}(2022)]%
        {wu2022pointtransformerv2}
\bibfield{author}{\bibinfo{person}{Xiaoyang Wu}, \bibinfo{person}{Yixing Lao}, \bibinfo{person}{Li Jiang}, \bibinfo{person}{Xihui Liu}, {and} \bibinfo{person}{Hengshuang Zhao}.} \bibinfo{year}{2022}\natexlab{}.
\newblock \showarticletitle{Point transformer v2: Grouped vector attention and partition-based pooling}.
\newblock \bibinfo{journal}{\emph{Advances in Neural Information Processing Systems (NeurIPS)}} (\bibinfo{year}{2022}).
\newblock


\bibitem[Wu et~al\mbox{.}(2024)]%
        {wu2024towards}
\bibfield{author}{\bibinfo{person}{Xiaoyang Wu}, \bibinfo{person}{Zhuotao Tian}, \bibinfo{person}{Xin Wen}, \bibinfo{person}{Bohao Peng}, \bibinfo{person}{Xihui Liu}, \bibinfo{person}{Kaicheng Yu}, {and} \bibinfo{person}{Hengshuang Zhao}.} \bibinfo{year}{2024}\natexlab{}.
\newblock \showarticletitle{Towards large-scale 3d representation learning with multi-dataset point prompt training}. In \bibinfo{booktitle}{\emph{Proceedings of the IEEE/CVF Conference on Computer Vision and Pattern Recognition (CVPR)}}. \bibinfo{publisher}{IEEE}, \bibinfo{pages}{19551--19562}.
\newblock


\bibitem[Wu et~al\mbox{.}(2023)]%
        {wu2023masked}
\bibfield{author}{\bibinfo{person}{Xiaoyang Wu}, \bibinfo{person}{Xin Wen}, \bibinfo{person}{Xihui Liu}, {and} \bibinfo{person}{Hengshuang Zhao}.} \bibinfo{year}{2023}\natexlab{}.
\newblock \showarticletitle{Masked scene contrast: A scalable framework for unsupervised 3d representation learning}. In \bibinfo{booktitle}{\emph{Proceedings of the IEEE/CVF Conference on Computer Vision and Pattern Recognition (CVPR)}}. \bibinfo{publisher}{IEEE}.
\newblock


\bibitem[Xie et~al\mbox{.}(2020)]%
        {xie2020pointcontrast}
\bibfield{author}{\bibinfo{person}{Saining Xie}, \bibinfo{person}{Jiatao Gu}, \bibinfo{person}{Demi Guo}, \bibinfo{person}{Charles~R Qi}, \bibinfo{person}{Leonidas Guibas}, {and} \bibinfo{person}{Or Litany}.} \bibinfo{year}{2020}\natexlab{}.
\newblock \showarticletitle{Pointcontrast: Unsupervised pre-training for 3d point cloud understanding}. In \bibinfo{booktitle}{\emph{European Conference on Computer Vision (ECCV)}}, Vol.~\bibinfo{volume}{12348}. \bibinfo{publisher}{Springer}, \bibinfo{pages}{574--591}.
\newblock


\bibitem[Yao et~al\mbox{.}(2024)]%
        {yao2024sgiformer}
\bibfield{author}{\bibinfo{person}{Lei Yao}, \bibinfo{person}{Yi Wang}, \bibinfo{person}{Moyun Liu}, {and} \bibinfo{person}{Lap-Pui Chau}.} \bibinfo{year}{2024}\natexlab{}.
\newblock \showarticletitle{SGIFormer: Semantic-guided and geometric-enhanced interleaving transformer for 3D instance segmentation}.
\newblock \bibinfo{journal}{\emph{IEEE Transactions on Circuits and Systems for Video Technology}} (\bibinfo{year}{2024}).
\newblock


\bibitem[Ye et~al\mbox{.}(2024)]%
        {gaussian_grouping}
\bibfield{author}{\bibinfo{person}{Mingqiao Ye}, \bibinfo{person}{Martin Danelljan}, \bibinfo{person}{Fisher Yu}, {and} \bibinfo{person}{Lei Ke}.} \bibinfo{year}{2024}\natexlab{}.
\newblock \showarticletitle{Gaussian Grouping: Segment and Edit Anything in 3D Scenes}. In \bibinfo{booktitle}{\emph{European Conference on Computer Vision (ECCV)}}, Vol.~\bibinfo{volume}{15087}. \bibinfo{publisher}{Springer}, \bibinfo{pages}{162--179}.
\newblock


\bibitem[Yeshwanth et~al\mbox{.}(2023)]%
        {yeshwanth2023scannet++}
\bibfield{author}{\bibinfo{person}{Chandan Yeshwanth}, \bibinfo{person}{Yueh-Cheng Liu}, \bibinfo{person}{Matthias Nie{\ss}ner}, {and} \bibinfo{person}{Angela Dai}.} \bibinfo{year}{2023}\natexlab{}.
\newblock \showarticletitle{Scannet++: A high-fidelity dataset of 3d indoor scenes}. In \bibinfo{booktitle}{\emph{Proceedings of the IEEE/CVF International Conference on Computer Vision (ICCV)}}. \bibinfo{publisher}{IEEE}.
\newblock


\bibitem[Yu and Song(2024)]%
        {yu2024mm}
\bibfield{author}{\bibinfo{person}{Hai-Tao Yu} {and} \bibinfo{person}{Mofei Song}.} \bibinfo{year}{2024}\natexlab{}.
\newblock \showarticletitle{Mm-point: Multi-view information-enhanced multi-modal self-supervised 3d point cloud understanding}. In \bibinfo{booktitle}{\emph{Proceedings of the AAAI Conference on Artificial Intelligence}}, Vol.~\bibinfo{volume}{38}. \bibinfo{pages}{6773--6781}.
\newblock


\bibitem[Yu et~al\mbox{.}(2022)]%
        {yu2022point}
\bibfield{author}{\bibinfo{person}{Xumin Yu}, \bibinfo{person}{Lulu Tang}, \bibinfo{person}{Yongming Rao}, \bibinfo{person}{Tiejun Huang}, \bibinfo{person}{Jie Zhou}, {and} \bibinfo{person}{Jiwen Lu}.} \bibinfo{year}{2022}\natexlab{}.
\newblock \showarticletitle{Point-bert: Pre-training 3d point cloud transformers with masked point modeling}. In \bibinfo{booktitle}{\emph{Proceedings of the IEEE/CVF Conference on Computer Vision and Pattern Recognition (CVPR)}}. \bibinfo{publisher}{IEEE}, \bibinfo{pages}{19291--19300}.
\newblock


\bibitem[Yue et~al\mbox{.}(2024)]%
        {yue2024improving}
\bibfield{author}{\bibinfo{person}{Yuanwen Yue}, \bibinfo{person}{Anurag Das}, \bibinfo{person}{Francis Engelmann}, \bibinfo{person}{Siyu Tang}, {and} \bibinfo{person}{Jan~Eric Lenssen}.} \bibinfo{year}{2024}\natexlab{}.
\newblock \showarticletitle{Improving 2d feature representations by 3d-aware fine-tuning}. In \bibinfo{booktitle}{\emph{European Conference on Computer Vision (ECCV)}}, Vol.~\bibinfo{volume}{15060}. \bibinfo{publisher}{Springer}, \bibinfo{pages}{57--74}.
\newblock


\bibitem[Zhao et~al\mbox{.}(2021)]%
        {zhao2021pointtransformer}
\bibfield{author}{\bibinfo{person}{Hengshuang Zhao}, \bibinfo{person}{Li Jiang}, \bibinfo{person}{Jiaya Jia}, \bibinfo{person}{Philip~HS Torr}, {and} \bibinfo{person}{Vladlen Koltun}.} \bibinfo{year}{2021}\natexlab{}.
\newblock \showarticletitle{Point transformer}. In \bibinfo{booktitle}{\emph{Proceedings of the IEEE/CVF International Conference on Computer Vision (ICCV)}}. \bibinfo{publisher}{IEEE}.
\newblock


\bibitem[Zheng et~al\mbox{.}(2025)]%
        {zheng2025survey}
\bibfield{author}{\bibinfo{person}{Ying Zheng}, \bibinfo{person}{Lei Yao}, \bibinfo{person}{Yuejiao Su}, \bibinfo{person}{Yi Zhang}, \bibinfo{person}{Yi Wang}, \bibinfo{person}{Sicheng Zhao}, \bibinfo{person}{Yiyi Zhang}, {and} \bibinfo{person}{Lap-Pui Chau}.} \bibinfo{year}{2025}\natexlab{}.
\newblock \showarticletitle{A survey of embodied learning for object-centric robotic manipulation}.
\newblock \bibinfo{journal}{\emph{Machine Intelligence Research}} (\bibinfo{year}{2025}), \bibinfo{pages}{1--39}.
\newblock


\bibitem[Zhou et~al\mbox{.}(2024)]%
        {zhou2024feature}
\bibfield{author}{\bibinfo{person}{Shijie Zhou}, \bibinfo{person}{Haoran Chang}, \bibinfo{person}{Sicheng Jiang}, \bibinfo{person}{Zhiwen Fan}, \bibinfo{person}{Zehao Zhu}, \bibinfo{person}{Dejia Xu}, \bibinfo{person}{Pradyumna Chari}, \bibinfo{person}{Suya You}, \bibinfo{person}{Zhangyang Wang}, {and} \bibinfo{person}{Achuta Kadambi}.} \bibinfo{year}{2024}\natexlab{}.
\newblock \showarticletitle{Feature 3dgs: Supercharging 3d gaussian splatting to enable distilled feature fields}. In \bibinfo{booktitle}{\emph{Proceedings of the IEEE/CVF Conference on Computer Vision and Pattern Recognition (CVPR)}}. \bibinfo{publisher}{IEEE}, \bibinfo{pages}{21676--21685}.
\newblock


\bibitem[Zhu et~al\mbox{.}(2023)]%
        {zhu2023ponderv2}
\bibfield{author}{\bibinfo{person}{Haoyi Zhu}, \bibinfo{person}{Honghui Yang}, \bibinfo{person}{Xiaoyang Wu}, \bibinfo{person}{Di Huang}, \bibinfo{person}{Sha Zhang}, \bibinfo{person}{Xianglong He}, \bibinfo{person}{Tong He}, \bibinfo{person}{Hengshuang Zhao}, \bibinfo{person}{Chunhua Shen}, \bibinfo{person}{Yu Qiao}, {et~al\mbox{.}}} \bibinfo{year}{2023}\natexlab{}.
\newblock \showarticletitle{Ponderv2: Pave the way for 3d foundataion model with a universal pre-training paradigm}.
\newblock \bibinfo{journal}{\emph{arXiv preprint arXiv:2310.08586}} (\bibinfo{year}{2023}).
\newblock


\bibitem[Zhu et~al\mbox{.}(2025)]%
        {zhu2025emosym}
\bibfield{author}{\bibinfo{person}{Yijie Zhu}, \bibinfo{person}{Yibo Lyu}, \bibinfo{person}{Zitong Yu}, \bibinfo{person}{Rui Shao}, \bibinfo{person}{Kaiyang Zhou}, {and} \bibinfo{person}{Liqiang Nie}.} \bibinfo{year}{2025}\natexlab{}.
\newblock \showarticletitle{EmoSym: A Symbiotic Framework for Unified Emotional Understanding and Generation via Latent Reasoning}. In \bibinfo{booktitle}{\emph{Proceedings of the 33nd ACM International Conference on Multimedia}}.
\newblock


\end{thebibliography}

\newpage
\appendix

\renewcommand{\thetable}{S.1}
\begin{table}[!htbp]
    \centering
    \caption{Implementation details of \ours.}
    \tablestyle{1.5pt}{0.9}
    \resizebox{0.45\textwidth}{!}{
    \begin{tabular}{y{30mm}x{25mm}}
        \toprule
        \textbf{Config} & \textbf{Value} \\
        \midrule
        \multicolumn{2}{c}{\textit{Training Details}} \\
        \midrule
        Optimizer & AdamW \\
        Betas & (0.9, 0.95) \\
        Weight Decay & 0.05 \\
        Learning Rate & 0.002 \\
        Learning Rate Scheduler & Cosine \\
        Batch Size & 32 \\
        Epochs & 1200 \\
        Warmup Epochs & 60 \\
        Mask Ratio & 50\% \\
        Masking Strategy & Random \\
        \midrule
        \multicolumn{2}{c}{\textit{Data Augmentation}} \\
        \midrule
        Random Rotation & $z$, $[-\pi, \pi]$, p: 1.0 \\
        {} & $x$, $[-\pi/64, \pi/64]$, p: 1.0 \\
        {} & $y$, $[-\pi/64, \pi/64]$, p: 1.0 \\
        Random Scaling & $[0.9, 1.1]$, p: 1.0 \\
        Random Flip & p : 0.5 \\
        Shuffle Point & p: 1.0 \\
        \bottomrule
    \end{tabular}
}
    \label{tab:sup_pretrain}
\end{table}

\section{Appendix Overview}

In this supplementary material, we provide more details about our proposed \ours. Specifically, we demonstrate more qualitative results, including visualization of learned representations, rendered images, depth maps, and semantic-aware feature maps. We also visualize the zero-shot representation of \ours on \sdis~\cite{armeni2016s3dis} and ScanNet++~\cite{yeshwanth2023scannet++}. In addition, we include implementation details for self-supervised representation learning and fine-tuning on downstream tasks.

\section{Qualitative Results}
\subsection{Visualization of Learned Representations}
In~\figref{fig:sup_teaser}, we visualize input point clouds, UMAP~\cite{mcinnes2018umap} results of learned representations, and corresponding synthetic RGB images, depth maps, and semantic-aware feature maps. From the results, we can observe that the learned point cloud representations are well clustered by UMAP on \scannet~\cite{dai2017scannet}, indicating that our \ours can effectively learn meaningful and expressive representations. For example, as shown in the second row, our learned representations are able to distinguish chairs and tables, proving that our model can reveal potential spatial relationships from the input point clouds by self-supervised learning. 

\renewcommand{\thetable}{S.2}
\begin{table}[!htbp]
    \centering
    \caption{Semantic segmentation settings of parameter efficiency~\cite{wu2025sonata}, data efficiency~\cite{hou2021csc}, and full fine-tuning on \scannet~\cite{dai2017scannet}, \scannetp~\cite{hou2021csc}, and \sdis~\cite{armeni2016s3dis}.}
    \tablestyle{1.5pt}{0.9}
    \resizebox{0.45\textwidth}{!}{
    \begin{tabular}{y{28mm}x{10mm}x{10mm}x{10mm}}
        \toprule
        \textbf{Config} & \multicolumn{3}{c}{\textbf{Value}} \\
        \cmidrule{2-4}
        {} & \scannet & \scannetp & \sdis \\
        \midrule
        Optimizer & \multicolumn{3}{c}{AdamW} \\
        Betas & \multicolumn{3}{c}{(0.9, 0.95)} \\
        Weight Decay & \multicolumn{3}{c}{0.05} \\
        Learning Rate & 0.005 & 0.005 & 0.003 \\
        Learning Rate Scheduler & \multicolumn{3}{c}{Cosine} \\
        Batch Size & 32 & 32 & 24 \\
        Data Efficiency Batch Size & 24 & - & - \\
        Epochs & 800 & 800 & 3000 \\
        Warmup Epochs & 40 & 40 & 150 \\
        Crop Size & 102400 & 102400 & 204800 \\
        Grid Sampling & \multicolumn{3}{c}{0.02m} \\
        \bottomrule
    \end{tabular}
}
    \label{tab:sup_sem}
\end{table}

\renewcommand{\thetable}{S.3}
\begin{table}[!htbp]
    \centering
    \caption{Instance segmentation settings on \scannet~\cite{dai2017scannet} and \scannetp~\cite{hou2021csc}.}
    \tablestyle{1.5pt}{0.9}
    \resizebox{0.45\textwidth}{!}{
    \begin{tabular}{y{28mm}x{12mm}x{12mm}}
        \toprule
        \textbf{Config} & \multicolumn{2}{c}{\textbf{Value}} \\
        \cmidrule{2-3}
        {} & \scannet & \scannetp \\
        \midrule
        Optimizer & \multicolumn{2}{c}{AdamW} \\
        Betas & \multicolumn{2}{c}{(0.9, 0.95)} \\
        Weight Decay & \multicolumn{2}{c}{0.05} \\
        Learning Rate & \multicolumn{2}{c}{0.005} \\
        Learning Rate Scheduler & \multicolumn{2}{c}{Cosine} \\
        Batch Size & 12 & 24 \\
        Epochs & \multicolumn{2}{c}{800} \\
        Warmup Epochs & \multicolumn{2}{c}{40} \\
        Crop Size & \multicolumn{2}{c}{Sample rate 0.8} \\
        Grid Sampling & \multicolumn{2}{c}{0.02m} \\
        \bottomrule
    \end{tabular}
}
    \label{tab:sup_ins}
\end{table}

\renewcommand{\thefigure}{S.1}
\begin{figure*}[!htbp]
    \centering
    \includegraphics[width=0.98\textwidth]{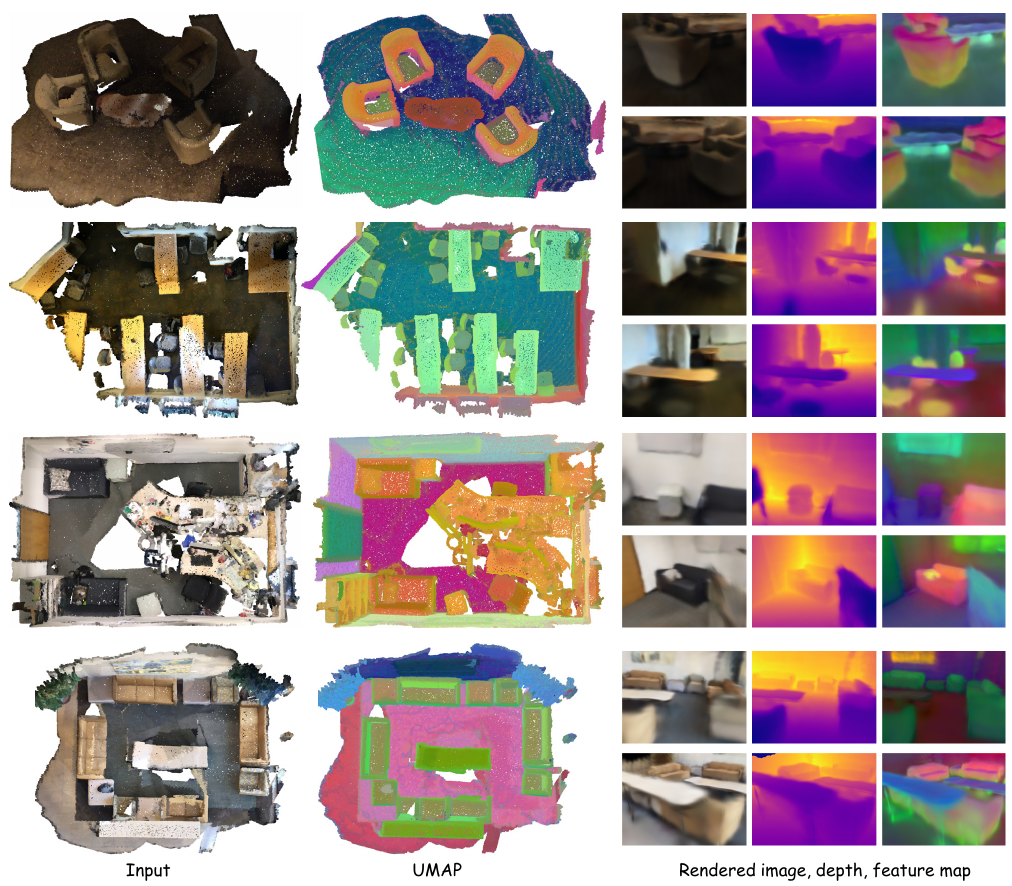}
    \caption{Qualitative results of \ours on \scannet~\cite{dai2017scannet}. We visualize the input point cloud and learned point representations using UMAP~\cite{mcinnes2018umap}. We also present the corresponding rendered images, depth maps, and semantic-aware feature maps.}
  \label{fig:sup_teaser}
\end{figure*}

\renewcommand{\thefigure}{S.2}
\begin{figure*}[!htbp]
    \centering
    \includegraphics[width=0.99\linewidth]{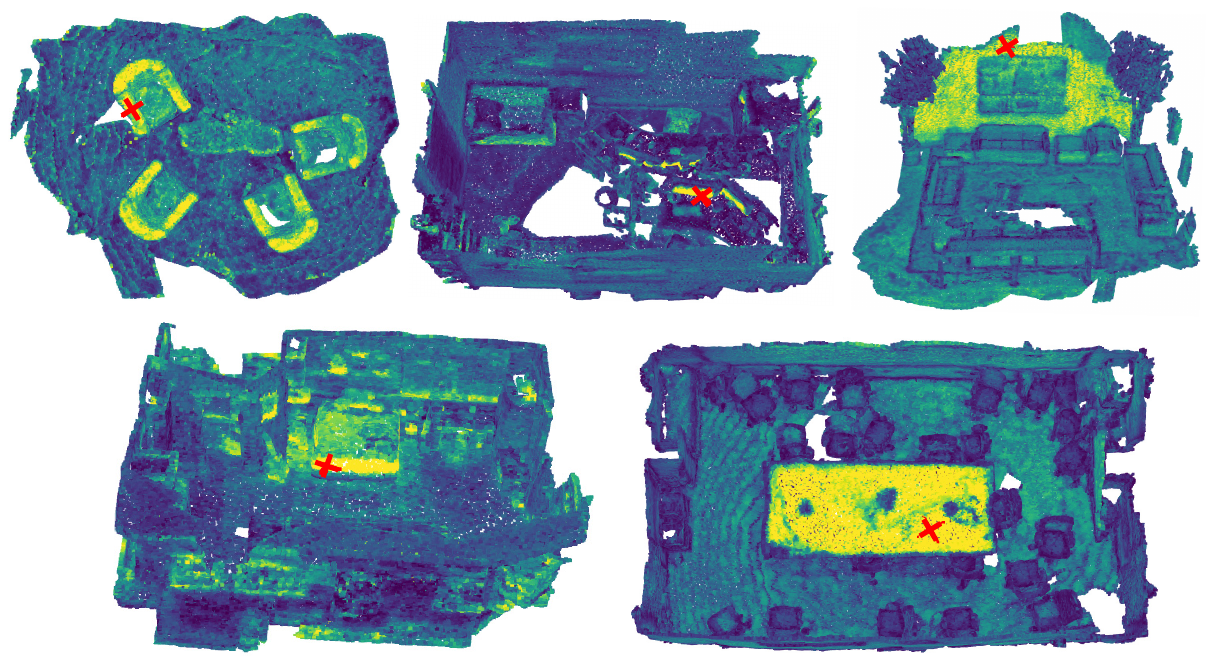}
    \caption{Visualization of activation maps of cosine similarity scores on \scannet~\cite{dai2017scannet}. The query points are highlighted with red cross marks.}
    \label{fig:sup_sim}
\end{figure*}

The color images, depth maps, and feature maps are rendered by our tri-attribute adaptive distillation splatting module during the pre-training process. Benefiting from the cuboid-normalized Gaussian initialization, our model can be generalizable to scale-variant point clouds. For instance, both the classroom (second row) and apartment (third row) scenes are well rendered with the correct colors and depth information. As for the semantic-aware feature maps, they can clearly recognize the semantic categories of the objects across different scenes, which is attributed to the incorporation of knowledge from 2D visual foundation models.

\subsection{Spatial Matching}
\subsubsection{In Domain Representation}
\label{subsubsec:in_domain_representation}

To further validate the quality of the learned representations by \ours, we visualize the dense spatial matching~\cite{wu2025sonata} results by some examples. Specifically, we select one query point from each scene and calculate the cosine similarity between the query point and others in the scene. We demonstrate the activation maps of the cosine similarity scores, where the brighter regions indicate higher similarity. The results on \scannet~\cite{dai2017scannet} are shown in~\figref{fig:sup_sim} with red cross marks highlighting the query points. We can observe that the learned representations are able to match the query points with their corresponding categories. For example, \ours can successfully match the query points of sofa, monitor, bed, table, and wall across scenes. This indicates that the model can learn discriminative representations, which is beneficial for downstream tasks such as semantic segmentation and instance segmentation.

\subsubsection{Zero-shot Representation} In~\figref{fig:sup_s3dis} and~\figref{fig:sup_scannetpp}, we visualize the zero-shot representation of \ours on \sdis~\cite{armeni2016s3dis} and ScanNet++~\cite{yeshwanth2023scannet++}. We directly apply the pre-trained weight on \scannet to these two unseen datasets without any fine-tuning and then visualize the results similar to~\figref{fig:sup_sim}. From the figures, we find that \ours demonstrates generalization ability to out-of-domain datasets.

\renewcommand{\thefigure}{S.3}
\begin{figure*}[!htbp]
    \centering
    \includegraphics[width=0.99\linewidth]{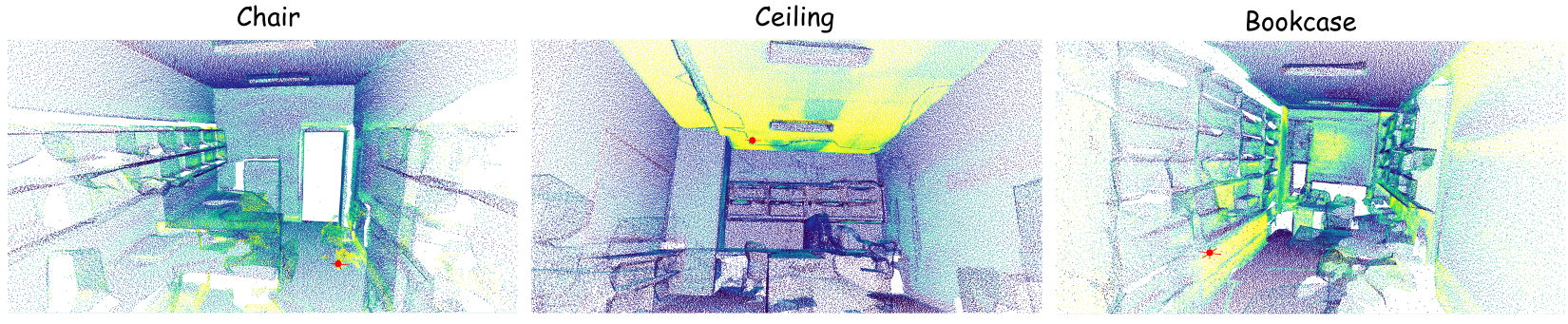}
    \caption{Zero-shot representation of \ours on \sdis~\cite{armeni2016s3dis}. The query points are highlighted with red circles.}
    \label{fig:sup_s3dis}
\end{figure*}

\renewcommand{\thefigure}{S.4}
\begin{figure*}[!htbp]
    \centering
    \includegraphics[width=0.99\linewidth]{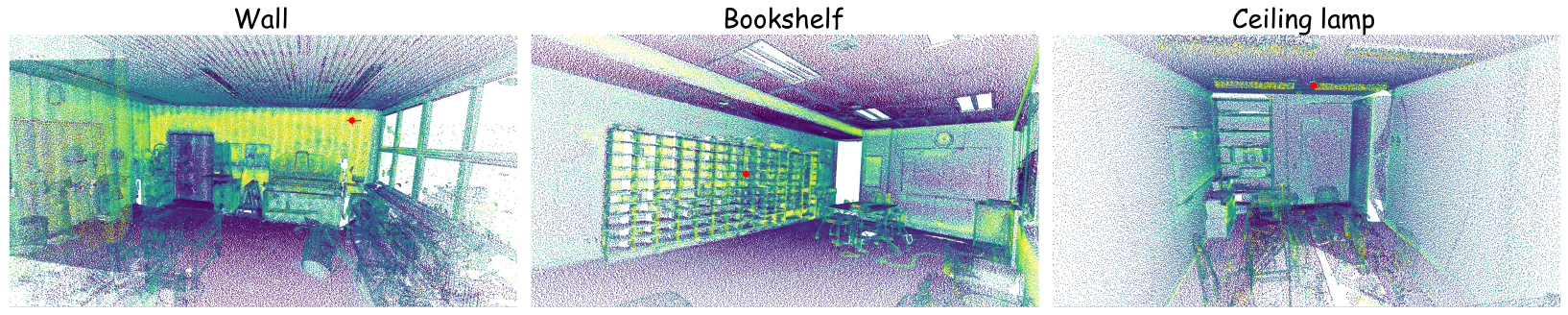}
    \caption{Zero-shot representation of \ours on ScanNet++~\cite{yeshwanth2023scannet++}. The query points are highlighted with red circles.}
    \label{fig:sup_scannetpp}
\end{figure*}

\subsection{Comparison with Ground Truth}
In~\figref{fig:sup_comparison}, we provide a qualitative comparison of \ours rendered images and depth maps with ground truth. We also show the synthesized semantic-aware feature maps. We can observe that the rendered images and depth maps are visually similar to the ground truth. Although there are some artifacts in the rendered images, the overall quality is still acceptable, and the rendered feature maps can help to alleviate this issue to some extent. Meanwhile, the depth information is also well-preserved to guarantee spatial consistency. This indicates that our tri-attribute adaptive distillation splatting can efficiently learn photometric appearance, geometrical structure, and semantic information simultaneously.

\renewcommand{\thefigure}{S.5}
\begin{figure*}[!htbp]
    \centering
    \includegraphics[width=0.99\linewidth]{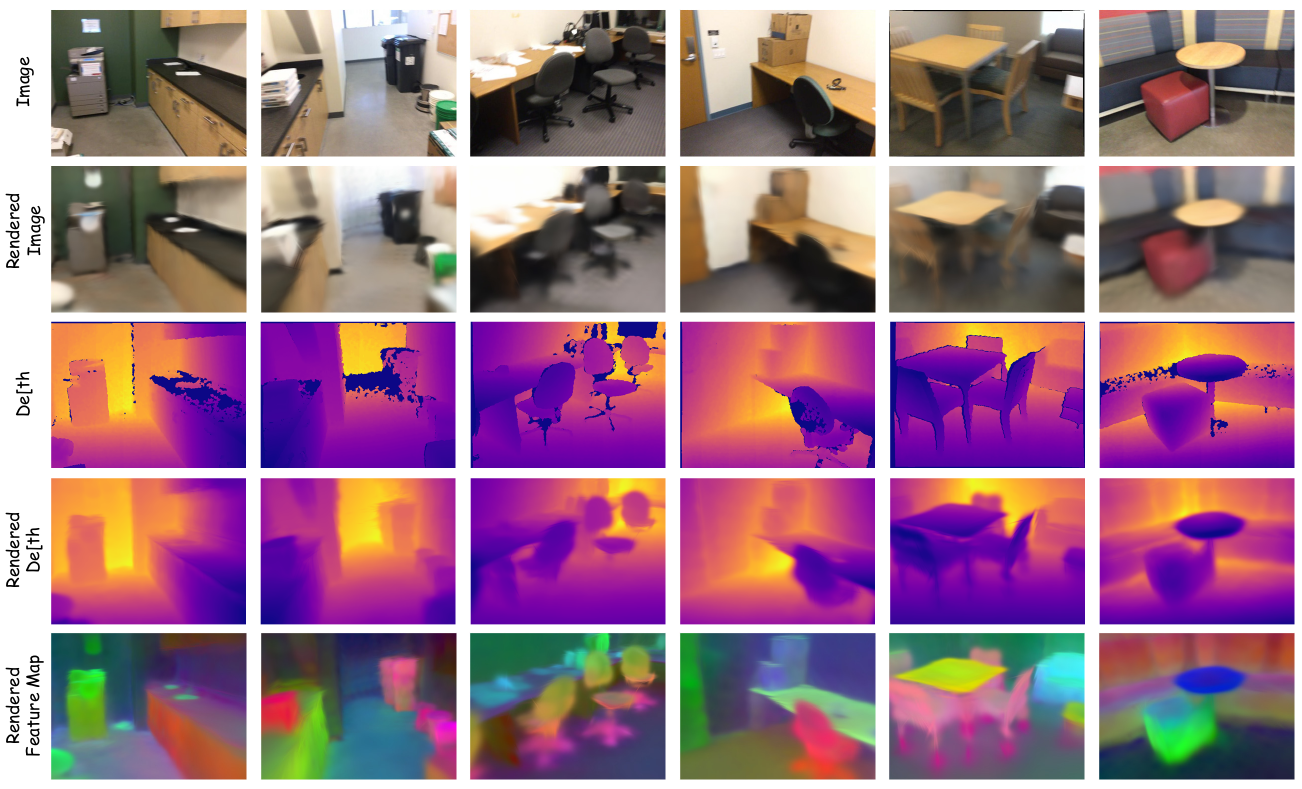}
    \caption{Qualitative comparison of \ours rendered images, depth, and semantic-aware feature maps with ground truth.}
    \label{fig:sup_comparison}
\end{figure*}

\section{Experimental Details}
\subsection{Pre-training}
We implement our \ours using Pointcept~\cite{pointcept2023} based on PyTorch. The self-supervised pre-training is conducted on \scannet~\cite{dai2017scannet}. The training details and data augmentations for the pre-training process are summarized in~\tabref{tab:sup_pretrain}. We adopt a 5-layer submanifold sparse convolutional U-Net~\cite{choy2019minkowski} (SparseUNet34C) as the point cloud backbone for performance comparison and ablation studies similar to MSC~\cite{wu2023masked}, PPT~\cite{wu2024towards}, and GC~\cite{wang2024groupcontrast}.

\subsection{Downsteam Tasks}
We use the same backbone architecture as the pre-training process for downstream tasks. The training details for semantic segmentation and instance segmentation are demonstrated in~\tabref{tab:sup_sem} and~\tabref{tab:sup_ins}, respectively. For parameter efficiency, data efficiency, and full fine-tuning, we follow the same settings. All downstream tasks are trained on 4 NVIDIA 4090 GPUs.

\end{document}